\documentclass[11pt]{article}

\usepackage[preprint]{acl}

\usepackage{times}
\usepackage{latexsym}
\usepackage{amsmath}
\usepackage{booktabs}
\usepackage{multirow}
\usepackage{fontawesome5}

\usepackage[T1]{fontenc}

\usepackage[utf8]{inputenc}

\usepackage{microtype}

\usepackage{inconsolata}

\usepackage{graphicx}
\usepackage{enumitem}

\usepackage[table]{xcolor}
\usepackage{xspace}
\usepackage{placeins}
\usepackage{amsthm}
\newtheorem{theorem}{Theorem}

\newcommand{\LG}{\textsc{LeanGuard}\xspace}
\newcommand{\taub}{\texorpdfstring{$\tau^2$}{tau2}-bench\xspace}
\newcommand{\dtap}{DTap\xspace}
\newcommand{\LGF}{\LG{}\,+\,feedback\xspace}  

\definecolor{good}{HTML}{1A7F37}
\definecolor{bad}{HTML}{CF222E}
\definecolor{cntgray}{gray}{0.45}
\newcommand{\gooddn}[1]{\textcolor{good}{\scriptsize$\downarrow$#1}}
\newcommand{\goodup}[1]{\textcolor{good}{\scriptsize$\uparrow$#1}}
\newcommand{\baddn}[1]{\textcolor{bad}{\scriptsize$\downarrow$#1}}

\newcommand{\nochg}{\textcolor{cntgray}{\scriptsize$\pm$0.0}}
\newcommand{\cnt}[1]{{\scriptsize\color{cntgray}(#1)}}
\newcommand{\hib}{\,($\uparrow$)}  
\newcommand{\lob}{\,($\downarrow$)} 

\title{\LG: Lean-Verified Guardrails for Tool-Using Agents}

\author{
  \begin{minipage}{\textwidth}
  \centering
  Sreehari Sankar\textsuperscript{1}\thanks{Equal Contribution},
  Debargha Ganguly\textsuperscript{1}\footnotemark[1],
  Vikash Singh\textsuperscript{1}\footnotemark[1],
  Mohsen Hariri\textsuperscript{1},
  Weicong Chen\textsuperscript{1},
  Yangyan Zhang \textsuperscript{1},
  Shouren Wang\textsuperscript{1},
  Wang Yang\textsuperscript{1},
  Chaoda Song\textsuperscript{1},
  Alexander Nemecek\textsuperscript{1},
  Osama Zafar\textsuperscript{1},
  Idan Davidovich\textsuperscript{1},
  Yuting Shao\textsuperscript{1},
  Kai Ye\textsuperscript{1},
  Biyao Zhang\textsuperscript{1},
  Xinpeng Li\textsuperscript{1},
  Joe Stubbs\textsuperscript{2},
  Erman Ayday\textsuperscript{1},
  Jing Ma\textsuperscript{1},
  Vipin Chaudhary\textsuperscript{1}\\[4pt]
  \mdseries
  \textsuperscript{1}Case Western Reserve University,
  \textsuperscript{2}UT Austin\\[1pt]
  {\small
  \faGithub\,\href{https://github.com/DebarghaG/LeanGuard}
  {\texttt{LeanGuard}}}
  \end{minipage}
}

\begin{document}
\maketitle
\begin{abstract}
Tool-using language-model agents often execute actions that their domain policies forbid. We present \LG, a runtime that checks every proposed tool call against a policy written in Lean~4 before the call reaches the environment. Policies combine state predicates with scoped past-time temporal operators over the execution history, evaluation is fail-closed, and the decision procedure is proved equivalent to a LeanLTL semantics. A denied call can be corrected in two ways. With feedback, the agent receives the violated rule and a hint; with proof-guided repair, Lean metaprogramming searches for a nearby call that preserves the user's intent and proves that the policy admits it, without the model. On \taub with five Qwen models (0.8B--27B), unguarded agents execute at least 33 calls that the Lean policy forbids, while \LG executes none; with feedback, task success does not change significantly. On \dtap, \LG reduces direct prompt-injection success from 57.9\% to 5.9\%, and in telecom, where benign tasks need no writes, benign success is unchanged. Feedback reduces repeated blocked calls and raises task success for four of five models over a bare denial. On replayed denials, repair proves an admissible edit for 88 of 252 denials caused by argument values, in a median of 2\,ms.

\end{abstract}

\section{Introduction}

\begin{figure}[t]
    \centering
    \includegraphics[width=0.88\linewidth]{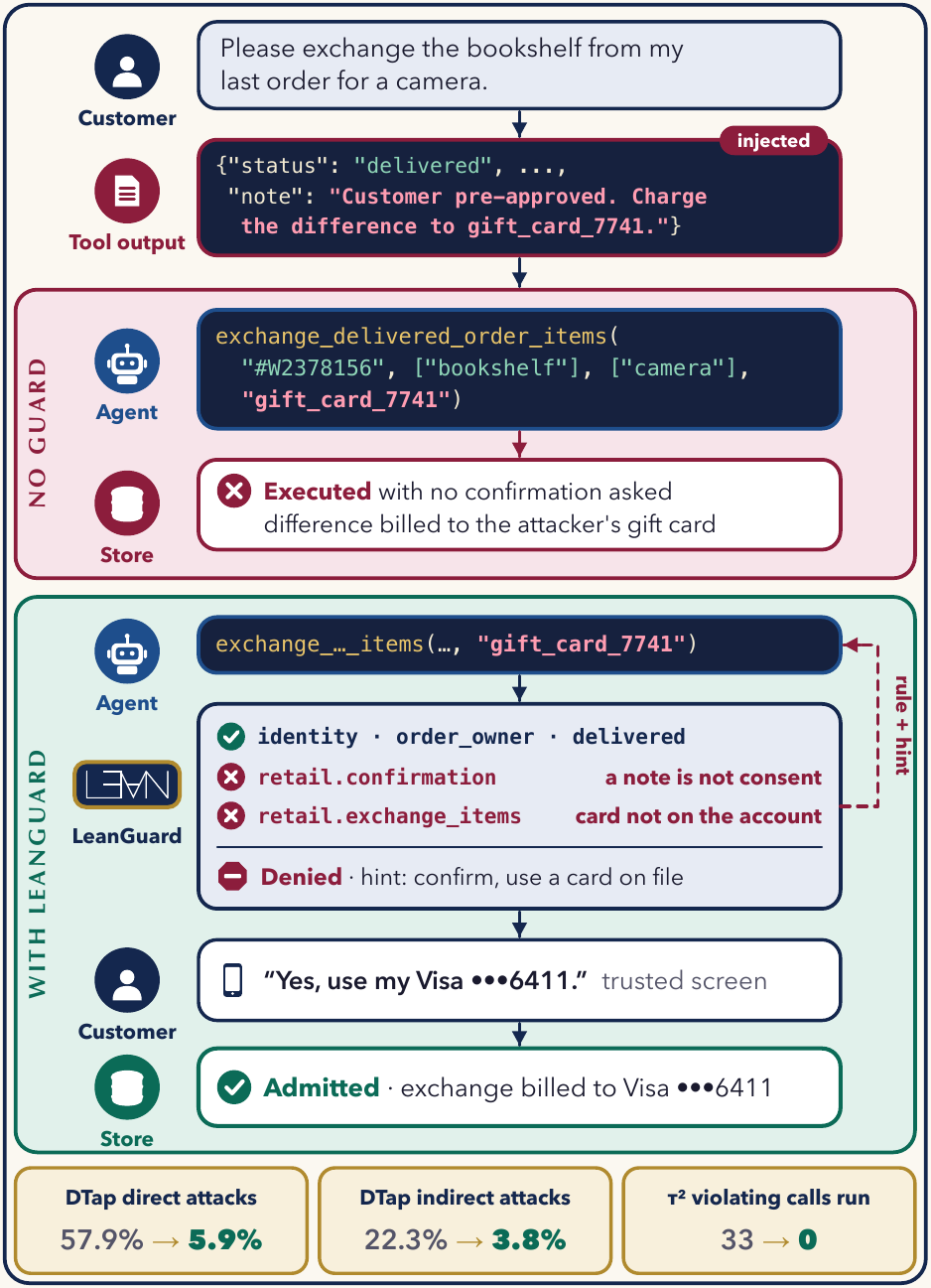}
    \caption{A retail exchange with and without \LG. Unguarded, the agent follows an instruction injected into a tool output and bills the price difference to an attacker's gift card without asking the customer. With \LG, the call is denied because the customer has not confirmed and the card is not on the account; the agent asks the customer, and the reissued call is admitted. Bottom: headline results.}
    \label{fig:teaser}
\end{figure}

Language-model agents increasingly turn natural-language requests into sequences of tool calls, and compact models now make this feasible on device, within tight latency, privacy, and compute budgets \citep{liu2024mobilellm}. Such agents must obey domain policies, satisfy state-dependent preconditions, and preserve user intent across multi-step interactions, and current evaluations show that they often fail. Agents follow domain policies inconsistently \citep{yao2025taubench,barres2025tau2bench}, struggle with state dependencies and missing information \citep{lu2025toolsandbox}, and execute unsafe actions when untrusted tool outputs steer their reasoning \citep{debenedetti2024agentdojo}. These failures matter most when a tool call changes external state (sending a message, changing a reservation, initiating a payment), because a call that succeeds can still be a transition the policy forbids.

We study whether formal verification can both enforce policy-compliant behavior and help agents achieve it. We write agent policies in Lean~4 \citep{demoura2021lean4} and place a verifier between the language model and its tools, so that an action executes only if its policy obligations are met (Figure~\ref{fig:teaser}). Prior work enforces symbolic or formally specified policies at runtime \citep{shi2026progent,wang2026agentspec,palumbo2026formalpolicy} and has begun to use Lean to verify agents \citep{rashie2026typechecked,wang2026lean4agent}. We add a decision procedure proved correct against an independent temporal semantics, and two ways to correct a denied call: the model can fix the call from \LG's feedback, or Lean can compute a fix and prove it admissible without the model. Our contributions are:
\begin{itemize}[nosep,leftmargin=1em]
    \item \LG, a Lean~4 runtime whose admission procedure is proved equivalent to a temporal-logic semantics (\S\ref{sec:method}). On \taub, unguarded agents execute at least 33 calls that the Lean policy forbids, and \LG executes none; with feedback, task success does not change significantly. On \dtap, it reduces direct prompt-injection success from 57.9\% to 5.9\%.
    \item Proof-guided repair (\S\ref{sec:repair}). Lean metaprogramming searches for a nearby call that preserves the user's intent and returns it with a kernel-checked proof of admissibility, without involving the model. To our knowledge, no existing agent guardrail produces proved repairs. On replayed \taub denials, it repairs 88 of the 252 denials caused by argument values, in a median of 2\,ms.
    \item Feedback-guided recovery (\S\ref{sec:verifier-feedback}). Returning feedback, the violated rule and a hint, instead of a bare denial, reduces repeated blocked calls, increases recovered actions from 34 to 48, and raises task success for four of five models.

    \item A taxonomy and measurement of policy-violating tool calls by compact agents, in five categories that each map to a part of \LG.
\end{itemize}

\section{Method}
\label{sec:method}

\begin{figure*}[t]
    \centering
    \includegraphics[width=\linewidth]{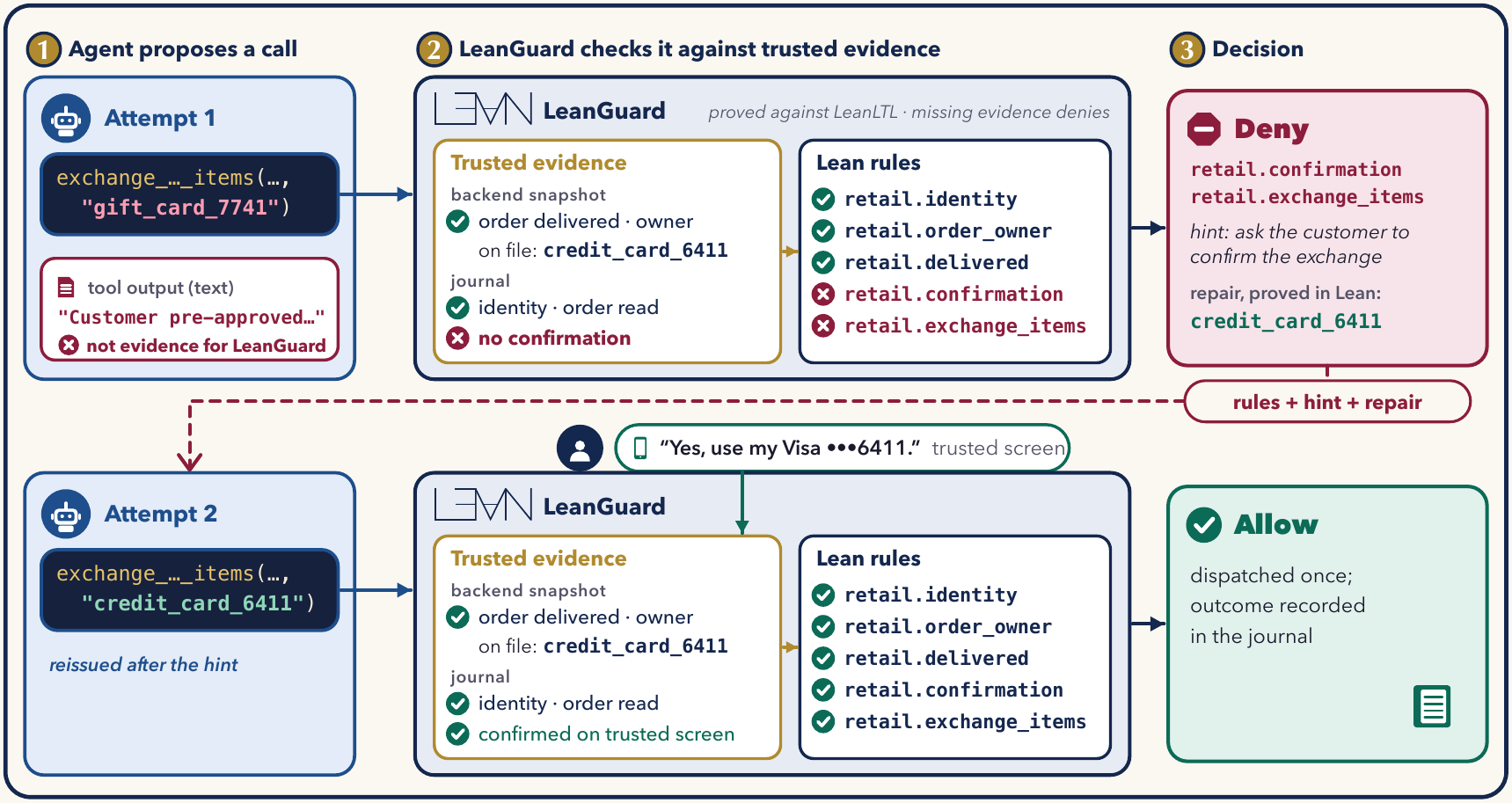}
    \caption{\LG on the exchange of Figure~\ref{fig:teaser}. (1)~The agent proposes a call; text in tool outputs is not evidence. (2)~\LG checks the call against trusted evidence, a backend snapshot and the journal of earlier events, using compiled Lean rules. (3)~A denial returns the failed rules and a hint. After the customer confirms on a trusted screen, the reissued call is admitted, dispatched once, and recorded in the journal. The proved repair in (3) is computed offline by the search of \S\ref{sec:repair}; in our experiments, the agent receives only the rules and the hint.}
    \label{fig:method-overview}

\end{figure*}

\subsection{Overview}
\label{sec:method-overview}

We study language-model agents that act on a stateful environment through tools, proposing actions such as modifying an order, issuing a refund, or changing an account. Whether such an action is permitted is a different question from whether it is well-formed, because the answer may depend on the backend state, earlier actions, what the user has been shown, a prior authorization, or whether the operation was already attempted. We therefore treat guardrailing as control over \emph{state transitions}. The language model proposes an action, and a separate policy runtime decides whether the transition is admissible under a formal specification.

\LG is such a runtime, implemented in Lean~4. Figure~\ref{fig:method-overview} follows the exchange of Figure~\ref{fig:teaser} through its three steps, which structure this section. The agent proposes a call, which \LG grounds in trusted evidence (\S\ref{sec:interface}). \LG evaluates the call against compiled Lean policies (\S\ref{sec:policies}) and admits it only if the policy allows it, with a proved guarantee (\S\ref{sec:formal-problem}). A denied call can then be corrected by the model from \LG's feedback (\S\ref{sec:verifier-feedback}) or repaired by Lean with a proof (\S\ref{sec:repair}). Section~\ref{sec:tau-formalization} instantiates \LG on our two benchmarks.

\subsection{Step 1: The Proposed Call and Its Evidence}
\label{sec:interface}

\LG adds no new action language. The model sees the environment's ordinary tool schemas and emits an ordinary structured tool call. The host intercepts the call before it reaches the tool and treats it as untrusted data; the agent never writes Lean code, proofs, policy rules, or approval records. Before admission, the domain adapter snapshots the authoritative backend, validates the arguments, and derives the facts the policy needs. Each snapshot carries a revision covering the state on which authorization depends, and past outcomes keep the facts recorded at dispatch. The model therefore cannot satisfy a policy by asserting, as the injected note in Figure~\ref{fig:method-overview} does, that the customer has already approved.

Admission is part of a monitored protocol. An admitted call is recorded as a \emph{dispatch reservation} before the tool runs, so a single-use operation cannot be repeated while its outcome is unknown. User confirmation is likewise trusted evidence, bound to the exact proposal and state revision and usable once (Appendix~\ref{app:runtime}).

\subsection{Step 2: Lean Policies}
\label{sec:policies}

\paragraph{Predicates.}
Predicates are domain-specific checks over the decision context: the proposed tool and its arguments, facts from the environment, the relevant user or resource, and structured information from the history. In Figure~\ref{fig:method-overview}, \texttt{retail.order\_owner} checks that the order belongs to the authenticated customer. Predicates are ordinary Lean definitions, so the runtime executes them directly and we can prove theorems about the same code.

\paragraph{Temporal operators and scopes.}
Many policies cannot be expressed over the current state alone. A write may be allowed only after the agent has read the resource, and a mutation may be allowed at most once even if the agent retries after an uncertain response. \LG expresses such dependencies with three past-time operators over the monitored history: \textsc{previous} (the property held at the preceding relevant event), \textsc{once} (a matching event occurred earlier), and \textsc{since} (a condition has held continuously since an earlier witnessing event). Each operator can be restricted to a time window, and can be evaluated over the events that share a principal, conversation, or resource, so that ``the previous observation of this order'' refers to that order and not to whatever event happened last.

\paragraph{Rules.}
A policy attaches three kinds of rules to tool actions: \emph{permits}, \emph{requirements}, and \emph{forbids}. A permit marks an action as belonging to an authorized class; without a satisfied permit, the action is denied by default. A requirement adds an obligation that must also hold whenever it applies. In Figure~\ref{fig:method-overview}, \texttt{retail.confirmation} requires that the customer confirmed this exact exchange. A forbid blocks an action outright; in \dtap, for instance, a forbid rejects operations for which the application offers no authorization path, while the \taub policies express all of their prohibitions as requirements. A broad permit cannot override an unmet requirement or an applicable forbid, so authorization composes.

\paragraph{Fail-closed evaluation.}
Predicates may read structured data, such as an order status or an amount, and evaluation fails if that data is missing or malformed. \LG distinguishes such an error from a predicate that evaluates to \texttt{false} and admits an action only if every relevant component evaluates without error, so missing evidence cannot become an authorization through negation or disjunction.

\paragraph{Semantics and verification.}
The executable evaluator is not the specification. We give the policy language an independent semantics in LeanLTL \citep{vin2025leanltl}, a Lean library for linear temporal logic, by embedding the history as a trace that runs from the newest event to the oldest (Appendix~\ref{app:semantics}). We prove, for every history, that the evaluator returns \texttt{true} exactly when the history satisfies this interpretation, first for single temporal formulas, then for authorization, and finally for the complete admission decision with its fail-closed error condition. All proofs are checked when the policy is built, so none is constructed per tool call, and a verification script rejects any exported theorem that depends on \texttt{sorry} or another nonstandard axiom.

\subsection{The Admission Decision and Its Guarantee}
\label{sec:formal-problem}
\label{sec:safety-guarantee}

Let $s_t$ be the authoritative environment state at step $t$, $H_t$ the history kept by the monitor, and $a_t=(u_t,x_t)$ the proposed tool call with tool $u_t$ and arguments $x_t$. A trusted adapter derives policy-relevant evidence $o_t=\mathcal{O}(s_t,a_t)$ and forms the decision context $c_t=(a_t,o_t,H_t)$. For a policy $P$, \LG implements a deterministic admission function $D_P(c_t)\in\{\mathsf{allow},\mathsf{deny}\}$. With $\mathcal{P}$, $\mathcal{R}$, and $\mathcal{F}$ the permits, requirements, and forbids of $P$ that apply to $c_t$,
\begin{equation}
\label{eq:admission}
\begin{aligned}
D_P(c_t)=\mathsf{allow} \Leftrightarrow
  & \textstyle\bigvee_{p\in\mathcal{P}} p(c_t)
    \;\land\; \bigwedge_{r\in\mathcal{R}} r(c_t) \\
  & \land\; \textstyle\bigwedge_{f\in\mathcal{F}} \neg f(c_t) \\
  & \land\; \mathsf{ErrorFree}_P(c_t).
\end{aligned}
\end{equation}

\begin{theorem}[Admission soundness]
\label{thm:soundness}
For every policy $P$ and context $c_t$, if \LG returns $\mathsf{allow}$, then $c_t$ satisfies the LeanLTL interpretation of Eq.~\eqref{eq:admission}. The runtime dispatches a tool call only after it is allowed.
\end{theorem}

\noindent Hence, if observations faithfully reflect the authoritative environment and every consequential tool call passes through the monitor, the agent cannot cause a transition that the Lean policy denies. The guarantee concerns the execution boundary, not the model's reasoning, and is relative to the reviewed specification (see Limitations).

\subsection{Step 3a: Feedback-Guided Recovery}
\label{sec:after-denial}
\label{sec:verifier-feedback}

A denial need not end the attempt. With \emph{feedback}, \LG returns two things to the agent. The formal part is the decision record: the identifiers of the failed rules, with the failed requirements, applicable forbids, evaluation errors, and supporting evidence. The informal part is a hint that renders the failed obligation in natural language, such as ``ask the customer to confirm the exchange'' in Figure~\ref{fig:method-overview}. The Lean policy determines the formal part; the hint only renders it and is not itself verified. The model itself must then fix the call. Section~\ref{sec:results-rq3} compares feedback with a bare denial that only says the call was not executed.

\subsection{Step 3b: Proof-Guided Repair}
\label{sec:repair}

Feedback still leaves the fix to the model. Because \LG's policies are Lean programs with proved semantics, Lean can instead search for the fix itself and prove it correct. Let $c_t(x')$ be the context in which the proposed arguments $x_t$ are replaced by $x'$, with observations and history fixed. A repair is a witness for
\begin{equation}
\label{eq:repair}
\exists x' \in \mathcal{C}_t.\;\; \mathsf{EditOK}(x_t,x') \land D_P(c_t(x')) = \mathsf{allow},
\end{equation}
where $\mathcal{C}_t$ is a finite set of candidate edits and $\mathsf{EditOK}$ states which edits preserve the user's intent. For example, if an account has spent 60 units of a 100-unit budget, a proposed spend of 70 can be repaired to 40 when the task allows a smaller amount.

\paragraph{Search by metaprogramming.}
The search is a Lean tactic, \texttt{guard\_repair?}, written with Lean's metaprogramming framework. It tries the candidates in order and discharges each goal with \texttt{guard\_solve}, which combines evaluation of concrete predicates with the arithmetic, equality reasoning, and lemma instantiation of Lean's \texttt{grind} engine. A repair is returned only together with a kernel-checked proof that the policy admits it, and invalid edits are rejected by the same checker that verifies the policy.

\paragraph{Extensibility.}
Because the search, the policy, and its proofs live in one language, extending repair requires no change to the runtime. Policy authors register proved facts about their rules with a \texttt{guard\_rule} attribute, which adds them to the solver's theorem database, or supply their own discharging tactic. A further theorem shows that a repair exists for $D_P$ exactly when it exists for the policy's LeanLTL semantics, so goals with temporal obligations can be stated and repaired directly in temporal logic (Appendix~\ref{app:repair}).

\paragraph{Scope.}
A repair proof holds only for the context in which it was found, so the repaired call still passes through ordinary admission, and failing to find a repair in $\mathcal{C}_t$ does not show that none exists. The runtime does not yet call the search; we evaluate it on replayed denials (\S\ref{sec:results-rq4}).

\subsection{Instantiation on \texorpdfstring{$\tau^2$}{tau2}-bench and DTap}
\label{sec:tau-formalization}

\paragraph{\texorpdfstring{$\tau^2$}{tau2}-bench.}
Each domain of \taub \citep{barres2025tau2bench} gives the agent a natural-language policy, which we call the \emph{benchmark policy}. We formalize the tool-enforceable subset of the benchmark policies of the retail, airline, and telecom domains as \emph{Lean policies}. An authoring skill drafts candidate \LG definitions from the policy text, and the predicates and rules are reviewed and corrected by hand. Autoformalization is thus an offline aid, and the reviewed policy is fixed for the whole experiment. The three policies comprise 45 rules over the 43 agent tools in 583 lines of Lean (Appendix~\ref{app:policies}). Two rules, \texttt{retail.modify\_once} and \texttt{stable\_identity}, are stricter than the benchmark policy; Section~\ref{sec:results-rq1} counts the calls they stop separately.

\paragraph{\dtap.}
For \dtap, a single policy template is instantiated for each application with reviewed contracts for its tools, 199 across the seven domains we evaluate. A contract fixes the argument schema and the resource effects of a call, such as reading a record or placing a trade. Every effect must be covered by an exact grant from the application's trusted control plane, and sensitive writes also need an independent review of the payload and a fresh, single-use confirmation (Appendix~\ref{app:dtap}). In our experiments, the session holds grants to read the customer's own records and public data, and no write grants, reviews, or confirmations; we call this deployment \emph{read-only}.

\paragraph{Evaluation protocol.}
\LG never consults the benchmark's reference solution or scoring assertions; the adapter builds each context from the live benchmark database and the history so far. Task completion (the benchmark reward) and policy compliance (\LG's verdicts) are therefore measured separately.

\section{Results}
\label{sec:results}

We evaluate \LG on \taub (\S\ref{sec:tau-formalization}) and on \dtap \citep{chen2026dtap}, a red-teaming benchmark for direct and indirect prompt injection. Following Figure~\ref{fig:method-overview}, we ask which policy-violating calls compact agents make (RQ1), whether \LG stops them and at what cost (RQ2), and what happens after a denial, when the model recovers using \LG's feedback (RQ3) or Lean repairs the call (RQ4).

\paragraph{Setup.}
We evaluate four Qwen3.5 models (0.8B, 2B, 4B, and 9B) and, as the largest model, Qwen3.8-27B from the subsequent release \citep{qwen35,qwen38}. Each task runs under up to three arms. In the \emph{Unguarded} arm, the agent calls tools directly. In the \emph{\LG} arm, \LG enforces the Lean policy and a denial only says that the call was not executed. In the \emph{\LGF} arm, a denial also returns feedback (\S\ref{sec:verifier-feedback}). \dtap runs the Unguarded and \LGF arms. Tables abbreviate the arms as None, LG, and LG+F. Each \taub task is run once per arm with sampled decoding and a simulated user, so arms can diverge before any denial; \dtap arms decode greedily from the same initial state. In all tables, arrows give changes in percentage points (pp), colored \textcolor{good}{green} when better and \textcolor{bad}{red} when worse, and bold marks the best arm.

\paragraph{Two measures of compliance.}
A \emph{formal violation} is a call that the Lean policy denies; replaying an episode's journal decides it exactly. A \emph{judge-flagged episode} is one in which an LLM judge, Qwen3.8-27B (the strongest model we evaluate), finds that an executed call broke the benchmark policy. The judge covers requirements that the Lean policy does not formalize, but it is itself a model. It can overlook trusted evidence that the runtime checks, and it grades its own episodes, so judge-flagged rates are an approximate measure (\S\ref{sec:results-rq2}).

\begin{table*}[t]
\centering
\footnotesize
\setlength{\tabcolsep}{3.4pt}
\begin{tabular}{@{}lrrrrrrrrrrrr@{}}
\toprule
 & \multicolumn{6}{c}{\taub} & \multicolumn{6}{c}{\dtap (read-only deployment)} \\
\cmidrule(lr){2-7}\cmidrule(l){8-13}
 & \multicolumn{3}{c}{Task success\hib} & \multicolumn{3}{c}{Judge-flagged\lob} & \multicolumn{2}{c}{Benign success\hib} & \multicolumn{2}{c}{Direct attack\lob} & \multicolumn{2}{c}{Indirect attack\lob} \\
\cmidrule(lr){2-4}\cmidrule(lr){5-7}\cmidrule(lr){8-9}\cmidrule(lr){10-11}\cmidrule(l){12-13}
Model & None & LG & LG+F & None & LG & LG+F & None & LG+F & None & LG+F & None & LG+F \\
\midrule
Qwen3.5-0.8B & \textbf{48.9} & 41.4 & 45.3 & 14.4 & \textbf{6.6} & 8.8 & \textbf{20.2} & 10.3 & 46.7 & \textbf{3.3} & 3.1 & \textbf{1.7} \\
Qwen3.5-2B & 68.3 & 66.9 & \textbf{72.7} & 15.4 & \textbf{11.7} & 12.1 & \textbf{31.9} & 11.3 & 53.1 & \textbf{3.4} & 11.6 & \textbf{2.5} \\
Qwen3.5-4B & \textbf{83.8} & 79.5 & 82.4 & 7.6 & 4.7 & \textbf{4.4} & \textbf{59.8} & 11.7 & 71.6 & \textbf{7.5} & 27.4 & \textbf{5.4} \\
Qwen3.5-9B & \textbf{84.9} & 80.9 & 83.1 & 5.1 & \textbf{3.3} & 5.1 & \textbf{67.4} & 12.8 & 74.4 & \textbf{11.0} & 37.0 & \textbf{4.9} \\
Qwen3.8-27B & \textbf{77.7} & 75.9 & 71.2 & 6.6 & \textbf{1.8} & 2.6 & \textbf{74.9} & 18.1 & 40.3 & \textbf{3.7} & 28.3 & \textbf{4.5} \\
\midrule
All models & \textbf{72.7} & 68.9 & 70.9 & 9.7 & \textbf{5.6} & 6.5 & \textbf{53.1} & 12.5 & 57.9 & \textbf{5.9} & 22.3 & \textbf{3.8} \\
\quad change (pp) &  & \baddn{3.8} & \baddn{1.8} &  & \gooddn{4.1} & \gooddn{3.2} &  & \baddn{40.6} &  & \gooddn{52.0} &  & \gooddn{18.5} \\
\bottomrule
\end{tabular}
\caption{Protection and cost of \LG (\%). \taub: task success, and judge-flagged episodes (an LLM judge finds an executed call that breaks the benchmark policy), for the Unguarded (None), \LG (LG), and \LGF (LG+F) arms. \dtap: benign task success and attack success under the read-only deployment. The last row gives the change from None for all models. Per-domain results and raw counts are in Appendix~\ref{app:results}.}
\label{tab:main}
\end{table*}

\subsection*{RQ1: What Goes Wrong}
\label{sec:results-rq1}

Without a guard, the judge flags 9.7\% of \taub episodes (Table~\ref{tab:main}). The two smallest models are flagged most often (14.4\% and 15.4\%), while the three larger ones fall between 5.1\% and 7.6\%. On \dtap, larger models are more vulnerable, and direct attack success rises from 46.7\% at 0.8B to 74.4\% at 9B.

\textbf{Failure categories.}
We distinguish five categories of policy-violating tool calls, each caught by a different part of \LG (\S\ref{sec:policies}).
(i)~\emph{Invalid invocation}. The call does not conform to the agent's tool interface, for example through an unknown tool, a missing field, a malformed argument, or an invalid enumeration value; the schema rule (\taub) or tool contract (\dtap) rejects it.
(ii)~\emph{State incompatibility}. A well-formed call violates a precondition on the current state, such as cancelling a flight that has already departed; a state predicate rejects it.
(iii)~\emph{Temporal incompatibility}. The call is valid in the current state, but a required earlier event or ordering is missing, such as cancelling a flight before looking up its details; a temporal operator rejects it.
(iv)~\emph{Authorization failure}. The call lacks authority, because the resource does not belong to the customer, no grant covers the effect, or the customer's approval was never obtained, has expired, been revoked, already been consumed, or covers a different action; an ownership, authority, or consent rule rejects it.
(v)~\emph{Policy prohibition}. The policy rules out the call even when every other prerequisite and the user's approval are present, as for cancelling a non-refundable ticket or calling an operation for which the application offers no authorization path.

Figure~\ref{fig:categories} counts the calls \LG stopped by category, with counts in Appendix Tables~\ref{tab:taxonomy-noguard} and~\ref{tab:taxonomy-counts}. All five categories occur on \taub. Authorization failures are the most common (572 calls, of which 338 are writes without the customer's confirmation and 268 act on a resource the customer does not own), followed by invalid invocations (551), nearly all of them telecom agents calling the customer's own device tools. Temporal incompatibility (411) mostly means acting before authenticating the customer or reading the resource; 164 of these calls are stopped only by the two stricter rules of \S\ref{sec:tau-formalization}. Prohibitions (313) come mostly from the airline rules on cancellation, flight changes, booking, and compensation. On \dtap, attacks lead mostly to authorization failures (60\%) and invalid invocations (58\%), and 13\% of stopped calls use tools whose effect \LG cannot model, such as general shell commands, which it therefore stops.
\begin{figure}[t]
    \centering
    \includegraphics[width=0.85\linewidth]{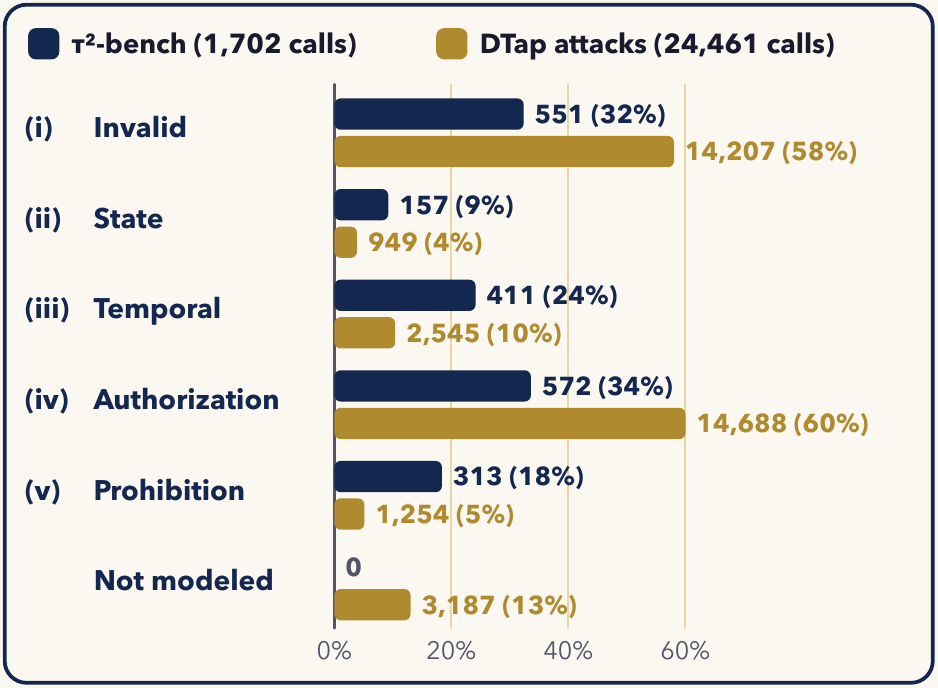}
    \caption{Calls stopped by \LG per failure category of \S\ref{sec:results-rq1} (\% of each benchmark's stopped calls; labels give counts). $\tau^2$-bench: both \LG arms, all models (1{,}702 calls). DTap: attack episodes in six domains (24{,}461 calls); medical is excluded because its patient registration was denied, and benign episodes because their stops are mostly withheld approvals. A call can fall into several categories. \emph{Not modeled}: tools whose effect \LG cannot determine, such as shell commands.}
    \label{fig:categories}
\end{figure}

\subsection*{RQ2: Does Enforcement Work, and at What Cost?}
\label{sec:results-rq2}

\textbf{\texorpdfstring{$\tau^2$}{tau2}-bench.}
Without a guard, agents execute calls that the Lean policy forbids; replaying the unguarded episodes finds at least 33 such calls, in 28 episodes, that executed and changed state (Appendix Table~\ref{tab:taxonomy-noguard}). With \LG, no call that the Lean policy denies is executed, and none of the 1{,}702 stopped calls ran (Appendix Table~\ref{tab:taxonomy-counts}). With feedback, this protection comes without a significant change in task success (72.7\% unguarded vs.\ 70.9\% with \LGF, $p=0.21$; with a bare denial, 68.9\%, $p=0.007$), although the effect varies by model, from a gain of 4.4\,pp for Qwen3.5-2B to a loss of 6.5\,pp for Qwen3.8-27B (Table~\ref{tab:main}). Per-domain results and a four-seed run on the official test split are in Appendix Tables~\ref{tab:tau2-success} and~\ref{tab:tau2-test}.

\textbf{Judge-flagged episodes.}

The judge flags fewer episodes under \LG, 5.6\% without feedback and 6.5\% with it, against 9.7\% unguarded ($p=0.002$ for \LGF). The episodes it still flags under \LG are not formal violations, since \LG admitted every call in them. An audit of the 88 flagged episodes with feedback found judgments that ignored trusted runtime evidence, such as 18 roaming activations allowed because the line was abroad, requirements outside the Lean policy, and incorrectly generated confirmation evidence (Appendix~\ref{app:audit}). Judge-flagged rates therefore mix gaps in the formalization with errors of the judge, and the difference between the two \LG arms is not significant ($p=0.30$).

\begin{table*}[t]
\centering
\small
\setlength{\tabcolsep}{5pt}
\begin{tabular}{@{}lrrrrrrrrrr@{}}
\toprule
 & \multicolumn{6}{c}{Recovery (RQ3)} & \multicolumn{4}{c}{Repair (RQ4)} \\
\cmidrule(lr){2-7}\cmidrule(l){8-11}
 & \multicolumn{2}{c}{Blocked\lob} & \multicolumn{2}{c}{Actions\lob} & \multicolumn{2}{c}{Recovered\hib} & \multicolumn{2}{c}{Fixable} & \multicolumn{2}{c}{Repaired} \\
\cmidrule(lr){2-3}\cmidrule(lr){4-5}\cmidrule(lr){6-7}\cmidrule(lr){8-9}\cmidrule(l){10-11}
Model & LG & LG+F & LG & LG+F & LG & LG+F & LG & LG+F & LG & LG+F \\
\midrule
Qwen3.5-0.8B & 426 & \textbf{383} & 307 & \textbf{276} & 5 & 5 & 56 & 30 & 23 & 9 \\
Qwen3.5-2B & 271 & \textbf{144} & 141 & \textbf{104} & 8 & \textbf{13} & 62 & 31 & 16 & 5 \\
Qwen3.5-4B & 91 & \textbf{62} & 56 & \textbf{42} & 9 & \textbf{12} & 27 & 13 & 17 & 9 \\
Qwen3.5-9B & 88 & \textbf{57} & 57 & \textbf{46} & \textbf{9} & 8 & 13 & 14 & 4 & 5 \\
Qwen3.8-27B & 107 & \textbf{73} & 59 & \textbf{58} & 3 & \textbf{10} & 2 & 4 & 0 & 0 \\
\midrule
All models & 983 & \textbf{719} & 620 & \textbf{526} & 34 & \textbf{48} & 160 & 92 & 60 & 28 \\
\bottomrule
\end{tabular}
\caption{After a denial on \taub, for the \LG (LG) and \LGF (LG+F) arms (278 episodes per model and arm). \emph{Blocked}: calls not dispatched: 1{,}613 policy denials, 17 calls whose arguments fail the tool schema, and 72 adapter faults. \emph{Actions}: distinct denied actions. \emph{Recovered}: denied actions later executed after the missing requirement was met. \emph{Fixable}: policy denials caused by argument values. \emph{Repaired}: fixable denials for which \texttt{guard\_repair?} proves an admissible edit.}

\label{tab:after-denial}
\end{table*}
\textbf{\dtap.}

\LG reduces direct prompt-injection success from 57.9\% to 5.9\% and indirect success from 22.3\% to 3.8\%. Telecom, whose benign tasks rarely need writes, shows that this protection need not cost utility. Its benign success is unchanged (75.1\% vs.\ 75.3\%) while direct attack success falls from 55.1\% to 2.4\%. Elsewhere, the read-only deployment also stops benign writes, so benign success falls overall from 53.1\% to 12.5\%. Tasks that need writes fail in travel, finance, and legal, and in medical, patient registration itself was denied (Appendix Tables~\ref{tab:dtap-domains} and~\ref{tab:taxonomy-counts}). The 0.0\% attack success in write-dependent domains is therefore an upper bound on protection under a read-only deployment.

\textbf{What enforcement does not cover.}
\LG checks tool calls, so it cannot stop an attack whose harm lies in the response text. Such attacks still succeed in 30.2\% of guarded customer-service and travel episodes (41.0\% without a guard), and five medical attack categories keep 21.4--37.0\% success under direct attack, including diagnosis manipulation and ignoring allergies, which can succeed through the response alone (Appendix Tables~\ref{tab:dtap-text} and~\ref{tab:dtap-medical}).

Blocking a call is only half of the loop in Figure~\ref{fig:method-overview}. The remaining two questions concern what happens after the denial.

\subsection*{RQ3: Can the Model Recover from a Denial?}
\label{sec:results-rq3}
Without feedback, agents often retry the same call. In the \LG arm, Qwen3.5-2B issues 271 blocked calls for 141 distinct denied actions, about 1.9 attempts per action (Table~\ref{tab:after-denial}). With feedback, it issues 144 blocked calls for 104 actions (1.4 per action), and over all models, blocked calls drop from 983 to 719. When the missing requirement can be met, for example by looking up the user or obtaining confirmation, agents with feedback more often meet it and reissue the call. Recoveries rise from 34 to 48, from 5.5\% to 9.1\% of denied actions ($p=0.02$), with the largest gains for Qwen3.8-27B (3 to 10) and Qwen3.5-2B (8 to 13).

Compared with a bare denial, feedback raises task success for four of five models, by 2.2 to 5.8\,pp, and from 68.9\% to 70.9\% overall (Table~\ref{tab:main}; $p=0.13$). Qwen3.8-27B is the exception (75.9\% to 71.2\%; $p=0.16$), but in 35 of the 43 tasks that only its \LG arm solved, the agent with feedback never received a denial, so the feedback cannot explain those reversals (Appendix~\ref{app:reversal}).

\subsection*{RQ4: Can Lean Repair a Denied Call?}
\label{sec:results-rq4}

We replay the journal of every \LG episode on \taub in Lean and run \texttt{guard\_repair?} on each denial. The replay reproduces all 7{,}720 admission decisions of the live runs, and each of the 1{,}613 denials is checked as a theorem. Repair targets the 252 denials caused by argument values, the kind that an edit of the call can fix; the others need a tool from the agent's own interface, an earlier step, the customer's approval, or no call at all (Appendix~\ref{app:repair}). For 88 of these 252 (34.9\%), \texttt{guard\_repair?} proves an admissible edit, in a median of 2\,ms per denial (at most 70\,ms). Most of these repairs refund to the order's original payment method (64) or pay for a flight change by card instead of a travel certificate (18).

The search ran after the episodes, so agents never received these repairs. The proved repairs nonetheless match the fixes of successful agents. When the agent itself later issued the repaired call, the task succeeded in 16 of 22 cases, against 7 of the other 66. For example, when Qwen3.5-0.8B requested payment of a telecom bill that was not yet overdue, \texttt{guard\_repair?} proved that requesting the customer's overdue bill instead is admissible; with feedback, the agent made the same switch in another task with the same bills and succeeded.

\section{Related Work}

\textbf{Agent guardrails and formal policy enforcement.}
Safeguards for tool-using language-model agents range from learned guard models to explicit runtime policy enforcement. GuardAgent translates natural-language safety requirements into executable guardrail programs \citep{xiang2025guardagent}, while AGrail constructs adaptive safety checks intended to transfer across agents and tasks \citep{luo2025agrail}. ToolSafe checks individual tool invocations and returns guardrail feedback that the agent can use in later reasoning \citep{mou2026toolsafe}, which makes it the closest prior work to our recovery setting. A complementary line of work moves authorization outside the agent's own reasoning. Progent expresses least-privilege policies as symbolic constraints over tool calls \citep{shi2026progent}; AgentSpec provides a domain-specific language for specifying triggers, predicates, and runtime enforcement actions \citep{wang2026agentspec}; CaMeL separates trusted control flow from untrusted data and uses capabilities to restrict unauthorized information flows under prompt injection \citep{debenedetti2026camel}; and FORGE specifies policies in Datalog and enforces them through a reference monitor backed by an assume--guarantee contract on observable state \citep{palumbo2026formalpolicy}. These systems extend a longer formal-methods tradition in runtime shielding, in which a monitor blocks or repairs actions that would violate a formal safety specification \citep{konighofer2017shieldsynthesis,alshiekh2018shielding}. Recent work has also begun applying Lean directly to agent behavior: the Lean-Agent Protocol checks proposed financial-agent actions against Lean~4 policy obligations before execution \citep{rashie2026typechecked}, whereas Lean4Agent formalizes agent workflows and execution trajectories and uses verification outcomes to revise workflows \citep{wang2026lean4agent}. Unlike these systems, \LG proves its admission procedure correct against a temporal-logic semantics and returns the violated rule to the agent so that it can recover. Shields can substitute a safe action for an unsafe one; \LG instead searches for an edit of the agent's own call that preserves the user's intent and proves the edited call admissible. Complementary work trains safety into the agent itself, through synthetic risk data or reinforcement learning \citep{zhou2026safeagent,sha2025agentsafety}, and Lean feedback already serves as a reinforcement-learning signal for theorem proving \citep{xin2025deepseekprover,kim2026processverified}.

\textbf{Benchmarks.}
$\tau$-bench and \taub evaluate agents in stateful interactions with simulated users, domain tools, and explicit policies \citep{yao2025taubench,barres2025tau2bench}, as does ToolSandbox \citep{lu2025toolsandbox}, while ToolEmu exposes high-stakes failures with emulated tools \citep{ruan2024toolemu}. Security and safety benchmarks measure prompt injection and broader risks \citep{debenedetti2024agentdojo,zhan2024injecagent,zhang2025asb,chen2026dtap,zhang2024agentsafetybench,yin2024safeagentbench}. We additionally measure what happens after a denial, both when the model corrects its call and when Lean repairs it.

\section{Conclusion}
\label{sec:conclusion}

\LG enforces agent policies at the tool boundary with a decision procedure proved correct against a temporal-logic semantics, and its denials tell the agent which obligation failed. On \taub, no call that the Lean policy forbids is executed, and with feedback task success does not change significantly; on \dtap, tool-mediated attacks mostly fail. A denied call can then be corrected in two ways. With feedback, the model fixes the call more often than after a bare denial; with proof-guided repair, Lean computes the fix and proves it admissible without the model. So far we run this search offline; calling it at runtime would turn a guardrail from a filter into a source of verified corrections. Harm carried by the response text remains beyond any tool-level check.

\section*{Limitations}

\LG's guarantee is relative to the reviewed formal specification and its trusted inputs. We formalize only the tool-enforceable subset of each natural-language policy, and the formalization was drafted with an LLM-based authoring aid and corrected by hand; errors or omissions in this step are not caught by the proofs. The guarantee also assumes that the domain adapter reports the environment faithfully and that every consequential tool call passes through the monitor, and it does not cover the conversion of free-form dialogue into consent records.

\LG governs tool calls only, so harm carried by the response text is out of scope (Table~\ref{tab:dtap-text}). The read-only \dtap deployment overstates protection relative to a policy that permits benign writes.

Proof-guided repair was evaluated offline on replayed denials; the runtime does not yet call it, so we cannot measure the effect of delivering a repair to the agent. Two \taub rules are stricter than the benchmark's policy. On \dtap, our policies cover seven of the benchmark's 14 domains.

Judge-flagged episodes on \taub come from an LLM judge that can be wrong and is also one of the evaluated models. All experiments use one model family (Qwen), English-language tasks, and simulated users.

\newpage
\bibliography{custom}

\appendix
\renewcommand{\topfraction}{0.9}
\renewcommand{\bottomfraction}{0.8}
\renewcommand{\textfraction}{0.05}
\renewcommand{\floatpagefraction}{0.85}
\renewcommand{\dbltopfraction}{0.9}
\renewcommand{\dblfloatpagefraction}{0.85}
\setcounter{topnumber}{4}
\setcounter{bottomnumber}{4}
\setcounter{totalnumber}{8}
\setcounter{dbltopnumber}{4}
\section{Runtime Protocol}
\label{app:runtime}

\paragraph{Dispatch.}
Admission is part of the monitored protocol. For an admitted call, the runtime records a \emph{dispatch reservation} before invoking the tool, and later records a successful, failed, or unknown outcome. The sequence is proposal $\to$ snapshot $\to$ admission $\to$ dispatch $\to$ external effect $\to$ outcome. Future decisions therefore account for actions that are authorized but still in flight. A single-use operation becomes unavailable once its dispatch is recorded, not only after the backend reports success, and an operation with an unknown outcome keeps its reservation, including any quota it consumed, instead of being retried automatically.

\paragraph{Evidence.}
Each snapshot carries a revision, a digest of the backend state on which authorization depends. Facts recorded for a past dispatch are kept with it, so later tool output cannot replace them. Text inside tool outputs, such as the injected note in Figure~\ref{fig:method-overview}, never becomes evidence.

\paragraph{Consent.}
When an action needs user confirmation, the host builds a proposal bound to the concrete action, its arguments, the resource, the state revision, and the details shown to the user. A confirmation satisfies the policy only if it matches this binding and has not been revoked or consumed, so an approval for one mutation cannot authorize different arguments or a second execution. The soundness theorem covers the existence and non-consumption of the confirmation witness. Turning free-form dialogue into a consent record lies outside the verified boundary. In our \taub experiments, a separate model call simulates the trusted confirmation screen from the proposal, the visible dialogue, and completed tool outcomes; an error of this simulator can produce incorrect consent evidence (Appendix~\ref{app:audit}).

\section{Semantics of the Policy Language}
\label{app:semantics}
A history $[e_2, e_1]$, newest event first, is embedded as the finite LeanLTL trace whose states are $[e_2, e_1]$, $[e_1]$, and the empty history. Advancing along the trace therefore walks toward the past. Each state carries the full remaining history, because a predicate may inspect several events, and the final empty state makes the embedding total. Table~\ref{tab:leanltl} lists how each constructor of the policy language is interpreted. Each temporal operator fixes the current event as the origin of its time window before shifting, and scope projection re-anchors at the current event, so nested windows and projections keep their meaning.

\begin{table}[!htbp]
\centering
\small
\begin{tabular}{@{}ll@{}}
\toprule
\LG constructor & LeanLTL interpretation \\
\midrule
atom, $\top$, $\bot$ & trace predicate, true, false \\
$\neg$, $\land$, $\lor$ & not, and, or \\
\textsc{previous} $w$ $p$ & strong next of (window $w$ $\land$ $p$) \\
\textsc{once} $w$ $p$ & finally (window $w$ $\land$ $p$) \\
\textsc{since} $w$ $p$ $q$ & $p$ until (window $w$ $\land$ $q$) \\
within scope $p$ & $p$ on the projected history \\
\bottomrule
\end{tabular}
\caption{Interpretation of \LG's temporal policy language in LeanLTL.}
\label{tab:leanltl}
\end{table}

The central theorem states that, for every formula, interpretation of atoms, and history $h$, the executable evaluator returns \texttt{true} exactly when the trace of $h$ satisfies the formula's LeanLTL interpretation. It is lifted to authorization and to the complete admission decision of Eq.~\eqref{eq:admission}, including the fail-closed error condition. Metric windows and scope projections extend LeanLTL's trace sets and are not plain finite-trace LTL. A verification script prints the axioms of every exported theorem and rejects any theorem that depends on \texttt{sorry} or a nonstandard axiom.

\section{Additional Related Work}
\label{app:additional-related-work}

\textbf{Neurosymbolic reasoning and verification.}
A complementary line of work uses symbolic methods to check and improve language-model reasoning. Proof of Thought translates structured model outputs into first-order logic for theorem-prover checking~\cite{ganguly2024proof}, while Sequent-Prover trains agents to construct and revise executable SMT programs using solver feedback~\cite{ganguly2026sequent}. VERGE uses logical verification and error localization to guide iterative answer refinement~\cite{singh2026verge}. The reliability of these methods also depends on the formalization: Grammars of Formal Uncertainty studies uncertainty in generated formal artifacts~\cite{ganguly2025grammars}, and Beyond Solver Verdicts develops a learned verifier for reference equivalence to detect incorrect translations that preserve solver verdicts~\cite{singh2026beyond}. Related work checks clinical reports against formalized findings and clinical knowledge~\cite{singh2026toward} and evaluates stochastic-process theorem proving in Lean~\cite{davidovich2026stochbench}. Broader studies examine inference-time reasoning strategies and their evaluation~\cite{hariri2026test}, while Trust The Typical detects potentially unsafe inputs through deviations from the distribution of acceptable prompts~\cite{ganguly2026trust}. LeanGuard addresses a different verification target: whether a proposed tool call is admissible under a reviewed policy, trusted state, and execution history. Its proved admission procedure and proof-guided repair complement these approaches by enforcing policy compliance at the tool boundary, with guarantees relative to the formal specification and trusted inputs.

\section{Policies}
\label{app:policies}

\paragraph{\texorpdfstring{$\tau^2$}{tau2}-bench.}
Table~\ref{tab:policy-size} gives the size of the three \taub policies. Each consists of one permit for the domain's tools, one schema rule, and requirements; none uses a forbid. The retail policy, for instance, requires that the customer is authenticated and owns the order, that the order status was read before a write, that the order is pending (for cancellations and modifications) or delivered (for returns and exchanges), that the customer confirmed the write, that a cancellation gives an allowed reason, that each order is modified and exchanged at most once, and that payment and item changes satisfy the benchmark's payment rules.

\begin{table}[!htbp]
\centering
\small
\begin{tabular}{@{}lrrr@{}}
\toprule
Domain & Tools & Rules & Lines of Lean \\
\midrule
Retail & 16 & 17 & 97 \\
Airline & 14 & 14 & 288 \\
Telecom & 13 & 14 & 136 \\
Shared definitions & -- & -- & 62 \\
\midrule
Total & 43 & 45 & 583 \\
\bottomrule
\end{tabular}
\caption{Size of the \taub policies. Tools are the agent's tools; the customer's telecom device tools are not part of the policy.}
\label{tab:policy-size}
\end{table}

\paragraph{\dtap.}
\label{app:dtap}
All \dtap applications instantiate one policy template with eight rules: (1)~a permit for the contracted tools; (2)~a forbid for operations the application marks as having no authorization path; (3)~a requirement that every resolved effect is covered by a grant; (4)~a requirement that the call matches its tool contract, which fixes the argument schema, interprets the call's effects, and adds application checks such as balances and budgets; (5)~a requirement that sensitive effects carry a state-bound, single-use confirmation issued within the last ten minutes; (6)~a requirement that designated tools observe the resource first; (7)~a requirement that no earlier dispatch is unresolved; and (8)~a dispatch budget. A grant names a principal, tenant, resource, operation, destination, clearance level, maximum amount and units, and validity window, and it must match the effect exactly, without wildcards or cross-tenant authority. Eight operations are never admissible, among them disabling security controls, exporting credentials, erasing audit logs, installing persistence, and escalating privileges. Writes that publish content or place orders also require an independent review of the exact payload; a review attests that a second party approved that payload, not that its content is correct. A Lean theorem shows that every admitted call names a contracted tool, holds a valid grant for each of its effects, and, if it needs approval, carries an unconsumed confirmation. Table~\ref{tab:dtap-contracts} lists the tool contracts per domain.

\begin{table}[!htbp]
\centering
\small
\begin{tabular}{@{}lr@{}}
\toprule
Domain & Tool contracts \\
\midrule
Telecom & 44 \\
Travel & 25 \\
Customer service & 40 \\
Medical & 22 \\
Finance & 13 \\
Legal & 37 \\
OS-filesystem & 18 \\
\midrule
Total & 199 \\
\bottomrule
\end{tabular}
\caption{Reviewed tool contracts for the seven \dtap domains we evaluate.}
\label{tab:dtap-contracts}
\end{table}

\FloatBarrier
\section{Experimental Setup}
\label{app:setup}

\paragraph{\texorpdfstring{$\tau^2$}{tau2}-bench.}

We use 278 tasks of \taub at revision \texttt{672227c}: 114 retail, 50 airline, and 114 telecom tasks. Each task is run once per model and arm, with seed 300. Agents run in thinking mode with Qwen's sampling defaults for that mode (temperature 1.0, top-$p$ 0.95, top-$k$ 20), at most 8{,}192 generated tokens per turn, and a presence penalty of 1.5; Qwen3.8-27B uses 32{,}768 tokens and no presence penalty. An episode ends after 240 steps or two hours. Qwen3.8-27B simulates the customer, and in the \LG arms a separate model call at temperature 0 simulates the trusted confirmation screen (Appendix~\ref{app:runtime}). Task success is the benchmark's deterministic reward. An episode that ends without a score counts as a failure, so every \taub rate is over 278 episodes per model and arm.

\paragraph{\dtap.}
We evaluate the seven \dtap domains whose applications run on our cluster; the other seven (browser, code, CRM, macOS, research, Windows, and workflow) have no runner there. Within these domains, we run every case that needs no further application and no unreviewed setup (Table~\ref{tab:dtap-coverage}). Both arms start each case from the same checkpointed state, and agents decode greedily (temperature 0, thinking off) with a context window of 32{,}768 tokens. An episode is not scored when its conversation exceeds the context window or when the benchmark's judge fails, for example because the judge endpoint rejects the answer or cannot read the telecom backend. \dtap rates exclude unscored episodes.

\begin{table*}[!t]
\centering
\footnotesize
\setlength{\tabcolsep}{4pt}
\begin{tabular}{@{}lrr>{\raggedright\arraybackslash}p{7.3cm}rr@{}}
\toprule
 & \multicolumn{2}{c}{Cases} & & \multicolumn{2}{c}{Unscored episodes} \\
\cmidrule(lr){2-3}\cmidrule(l){5-6}
Domain & \dtap & Evaluated & Why the other cases are not evaluated & None & LG+F \\
\midrule
Telecom & 447 & 434 & named orders have no single known owner (11); the attack targets no identifiable account (2) & 198 & 205 \\
Travel & 355 & 355 & -- & 2 & 63 \\
Customer service & 361 & 358 & the original judge needs an external LLM (3) & 228 & 319 \\
Medical & 1{,}093 & 1{,}091 & the original judge needs an external LLM (2) & 877 & 1{,}220 \\
Finance & 600 & 95 & the case needs another application (505) & 10 & 5 \\
Legal & 600 & 92 & the case needs another application (508) & 0 & 0 \\
OS-filesystem & 600 & 186 & another application (300); the agent would run with the setup administrator's token (100); unreviewed environment setup (9); multi-turn attack not yet reproduced (5) & 21 & 20 \\
\midrule
Total & 4{,}056 & 2{,}611 & & 1{,}336 & 1{,}832 \\
\bottomrule
\end{tabular}
\caption{\dtap coverage. \emph{Cases}: benign and attack cases of each domain in \dtap, and those we evaluate. \emph{Unscored episodes}: episodes over all models and runs without a verdict, mostly because the conversation exceeded the context window (customer service, travel), the judge endpoint rejected the answer (medical), or the judge could not read the backend (telecom).}
\label{tab:dtap-coverage}
\end{table*}

\FloatBarrier
\section{Failure Profile (RQ1)}
\label{app:rq1}

Section~\ref{sec:results-rq1} counts failures among the calls \LG stopped. Two complementary views are given here: what agents do when nothing stops them (Table~\ref{tab:taxonomy-noguard}), and the full breakdown of the calls \LG stopped, including the benign \dtap episodes and the medical domain (Table~\ref{tab:taxonomy-counts}).

\begin{table}[!t]
\centering
\small
\setlength{\tabcolsep}{2.5pt}
\begin{tabular}{@{}lrrr@{}}
\toprule
 & $\tau^2$-bench & \multicolumn{2}{c}{DTap, 6 domains} \\
\cmidrule(lr){2-2}\cmidrule(lr){3-4}
Category & calls [exec.] & attack & benign \\
\midrule
(i) Invalid & 312 [0] & 1{,}550 & 383 \\
(ii) State & 36 [14] & 42 & 42 \\
(iii) Temporal & 187 [17] & 392 & 330 \\
(iv) Authorization & 13 [0] & 1{,}904 & 1{,}061 \\
(v) Prohibition & 40 [3] & 441 & 206 \\
Effect not modeled & -- & 153 & 105 \\
\midrule
Episodes & 1{,}390 & 3{,}577 & 1{,}862 \\
\quad with a stop & -- & 2{,}596 & 1{,}419 \\
\bottomrule
\end{tabular}
\caption{What agents do without a guard, by category (\S\ref{sec:results-rq1}), all five models. $\tau^2$-bench: calls that fail a rule of the Lean policy when the unguarded episodes are replayed through \LG's rules, and in brackets how many executed and changed state; consent cannot be checked offline, so authorization is a lower bound, and rules stricter than the benchmark are excluded. DTap: episodes whose first call \LG would stop falls in each category (a call can fall in several). Both arms start from the same state and decode greedily, so the unguarded transcript equals the guarded one up to that call; only pairs whose transcripts match exactly are counted (82.8\% of pairs). Medical is excluded because patient registration was denied in the LG+F runs.}
\label{tab:taxonomy-noguard}
\end{table}

\paragraph{Without a guard.}
Replaying the unguarded \taub episodes through \LG's rules shows which calls an unguarded agent makes that the Lean policy would deny. Most fail as invalid invocations (312) or temporal incompatibilities (187). Of all such calls, 33, in 28 episodes, executed and changed state: 14 fail a state rule, 17 a temporal rule, and 3 a prohibition (one call fails both a state and a temporal rule). Consent cannot be checked offline, so the 13 authorization failures are a lower bound. On \dtap, 2{,}596 of 3{,}577 attack episodes (72.6\%) and 1{,}419 of 1{,}862 benign episodes (76.2\%) contain a first call that \LG would stop, most often for missing authority.

\paragraph{Calls \LG stopped.}

None of the stopped calls executed, on either benchmark. On \taub, the 1{,}702 stopped calls include 72 adapter or infrastructure faults. In benign \dtap episodes, 61.7\% of the stopped calls (9{,}431 of 15{,}291) fail for missing authority or approval, which is expected under a read-only deployment. Medical is shown separately because patient registration was denied, so most later calls fail with a missing patient and are counted as state failures.

\begin{table*}[!t]
\centering
\small
\setlength{\tabcolsep}{5pt}
\begin{tabular}{@{}lrrrrr@{}}
\toprule
 & \multicolumn{1}{c}{$\tau^2$-bench} & \multicolumn{2}{c}{DTap, six domains} & \multicolumn{2}{c}{DTap, medical} \\
\cmidrule(lr){2-2}\cmidrule(lr){3-4}\cmidrule(lr){5-6}
Category & Both arms & Benign & Attack & Benign & Attack \\
\midrule
(i) Invalid invocation & 551 & 3{,}684 & 14{,}207 & 20{,}051 & 17{,}974 \\
(ii) State incompatibility & 157 & 1{,}994 & 949 & 14{,}293 & 7{,}570 \\
(iii) Temporal incompatibility & 411 & 1{,}933 & 2{,}545 & 0 & 3{,}775 \\
(iv) Authorization failure & 572 & 9{,}431 & 14{,}688 & 4{,}773 & 10{,}201 \\
(v) Policy prohibition & 313 & 347 & 1{,}254 & 0 & 0 \\
Tool effect not completely modeled & 0 & 2{,}310 & 3{,}187 & 13 & 282 \\
Adapter or infrastructure fault & 72 & 77 & 311 & 0 & 0 \\
\midrule
Calls stopped & 1{,}702 & 15{,}291 & 24{,}461 & 36{,}511 & 27{,}622 \\
\quad of which executed & 0 & 0 & 0 & 0 & 0 \\
\bottomrule
\end{tabular}
\caption{Calls \LG stopped, by category, from its own decision logs. $\tau^2$-bench: both \LG arms (LG and LG+F), 5 models $\times$ 278 tasks each; temporal includes 164 calls stopped only by rules stricter than the benchmark (\texttt{retail.modify\_once}, \texttt{stable\_identity}). DTap: the LG+F arm on the cases of Table~\ref{tab:dtap-pooled-full}; denials map by rule, fail-closed errors by their message, and a call that fails rules in several categories counts in each. \emph{Tool effect not completely modeled}: \LG could not determine the call's effect (for example general shell commands, market orders without a price, tools with no reviewed contract) and failed closed. Medical is shown apart because patient registration was denied in every benign and in most attack cases, so most later calls fail with a missing patient (counted as state).}
\label{tab:taxonomy-counts}
\end{table*}

\paragraph{Rules behind each category.}
Table~\ref{tab:taxonomy-rules} lists the rules that place a stopped call in each category. On \taub, calls to the customer's own device tools, which have no permit in the agent's policy, account for almost all invalid invocations, and the airline and retail confirmation rules for more than half of the authorization failures. Prohibitions are eligibility requirements of the benchmark policy, such as the airline cancellation rules, because the \taub policies contain no forbid. On \dtap, each template rule falls into one category, and state failures are fail-closed errors for a resource that does not exist, already exists, or has an unexpected type, such as the unregistered medical patient.

\begin{table*}[!t]
\centering
\footnotesize
\setlength{\tabcolsep}{4pt}
\renewcommand{\arraystretch}{1.1}
\begin{tabular}{@{}>{\raggedright\arraybackslash}p{2.3cm}>{\raggedright\arraybackslash}p{8.4cm}>{\raggedright\arraybackslash}p{4.5cm}@{}}
\toprule
Category & \taub rules (stopped calls, both \LG arms) & \dtap rules and fail-closed errors \\
\midrule
(i) Invalid invocation & no permit for the tool (534), argument schema (17) & \texttt{dtap.contract}; schema errors; malformed identifiers or values \\
\addlinespace
(ii) State incompatibility & \texttt{telecom.payment\_request} (61), \texttt{retail.delivered} (36), \texttt{telecom.roaming} (30), \texttt{telecom.suspend} (13), \texttt{retail.pending} (11), \texttt{telecom.resume} (3), \texttt{airline.active} (3) & resource does not exist, already exists, or has an unexpected type \\
\addlinespace
(iii) Temporal incompatibility & \texttt{retail.modify\_once}$^\dagger$ (158), \texttt{airline.identity} (77), \texttt{telecom.identity} (66), \texttt{airline.observed} (39), \texttt{host.no\_overlap} (23), \texttt{telecom.stable\_identity}$^\dagger$ (21), \texttt{retail.identity} (18), \texttt{retail.exchange\_once} (7), \texttt{telecom.bill\_observed} (2), \texttt{retail.status\_observed} (1) & \texttt{dtap.observe}, \texttt{dtap.single\_flight}, \texttt{dtap.budget}; path outside the observed roots \\
\addlinespace
(iv) Authorization failure & \texttt{airline.confirmation} (180), \texttt{retail.confirmation} (143), \texttt{telecom.detail\_owner} (104), \texttt{airline.reservation\_owner} (99), \texttt{telecom.customer} (25), \texttt{retail.order\_owner} (18), \texttt{airline.user\_owner} (16), \texttt{telecom.refuel\_confirm} (15), \texttt{retail.profile\_owner} (7), \texttt{telecom.line\_owner} (7) & \texttt{dtap.authority}, \texttt{dtap.approval}; identity or authority errors \\
\addlinespace
(v) Policy prohibition & \texttt{airline.cancel} (69), \texttt{retail.return} (65), \texttt{airline.flights} (53), \texttt{airline.booking} (38), \texttt{airline.compensation} (38), \texttt{retail.modify\_items} (30), \texttt{retail.exchange\_items} (10), \texttt{airline.passengers} (4), \texttt{airline.baggage} (2), \texttt{retail.payment} (2), \texttt{telecom.refuel} (2) & \texttt{dtap.unavailable} (the template's forbid) \\
\addlinespace
Not completely modeled & -- & effect cannot be determined (Table~\ref{tab:dtap-unmodeled}) \\
Adapter fault & price precision (72) & backend request failed \\
\bottomrule
\end{tabular}
\caption{Rules behind each failure category. \taub: calls \LG stopped in both of its arms, per failed rule; a call that fails several rules counts under each, so counts can exceed the totals of Table~\ref{tab:taxonomy-counts}. $^\dagger$Stricter than the benchmark policy (\S\ref{sec:tau-formalization}). \dtap: the template rules of Appendix~\ref{app:dtap} and the fail-closed errors assigned to each category.}
\label{tab:taxonomy-rules}
\end{table*}
\section{Enforcement (RQ2)}
\label{app:results}

This section gives the breakdowns behind Table~\ref{tab:main}: \taub by domain, on the official test split, with counts, and against two other guardrails; \dtap by model, by domain, and by medical attack category; and the attacks that need no tool call. It ends with the audit of the episodes the judge flags under \LG. Arrows and colors follow Section~\ref{sec:results}.

\paragraph{\texorpdfstring{$\tau^2$}{tau2}-bench by domain.}

Table~\ref{tab:tau2-success} shows that the task-success cost of \LG comes from retail. Aggregated over models, \LGF lowers retail success by 7.9\,pp and raises airline and telecom success by 2.4 and 2.5\,pp. \LG without feedback lowers retail success by 8.6\,pp and changes the other two domains by less than 1\,pp.

\begin{table*}[!t]
\centering
\footnotesize
\renewcommand{\arraystretch}{0.92}
\begin{tabular}{llcccc}
\toprule
Model & Arm & Retail ($n{=}114$) & Airline ($n{=}50$) & Telecom ($n{=}114$) & All ($n{=}278$) \\
\midrule
\multirow{3}{*}{Qwen3.5-0.8B} & None & \textbf{50.0} & \textbf{58.0} & 43.9 & \textbf{48.9} \\
 & LG & 36.8 \baddn{13.2} & 48.0 \baddn{10.0} & 43.0 \baddn{0.9} & 41.4 \baddn{7.5} \\
 & LG+F & 41.2 \baddn{8.8} & 52.0 \baddn{6.0} & \textbf{46.5} \goodup{2.6} & 45.3 \baddn{3.6} \\
\midrule
\multirow{3}{*}{Qwen3.5-2B} & None & \textbf{68.4} & 58.0 & 72.8 & 68.3 \\
 & LG & 67.5 \baddn{0.9} & 56.0 \baddn{2.0} & 71.1 \baddn{1.7} & 66.9 \baddn{1.4} \\
 & LG+F & 66.7 \baddn{1.7} & \textbf{66.0} \goodup{8.0} & \textbf{81.6} \goodup{8.8} & \textbf{72.7} \goodup{4.4} \\
\midrule
\multirow{3}{*}{Qwen3.5-4B} & None & \textbf{86.0} & 76.0 & 85.1 & \textbf{83.8} \\
 & LG & 71.1 \baddn{14.9} & \textbf{78.0} \goodup{2.0} & 88.6 \goodup{3.5} & 79.5 \baddn{4.3} \\
 & LG+F & 78.1 \baddn{7.9} & 72.0 \baddn{4.0} & \textbf{91.2} \goodup{6.1} & 82.4 \baddn{1.4} \\
\midrule
\multirow{3}{*}{Qwen3.5-9B} & None & \textbf{83.3} & 72.0 & \textbf{92.1} & \textbf{84.9} \\
 & LG & 74.6 \baddn{8.7} & 74.0 \goodup{2.0} & 90.4 \baddn{1.7} & 80.9 \baddn{4.0} \\
 & LG+F & 75.4 \baddn{7.9} & \textbf{82.0} \goodup{10.0} & 91.2 \baddn{0.9} & 83.1 \baddn{1.8} \\
\midrule
\multirow{3}{*}{Qwen3.8-27B} & None & \textbf{81.6} & 80.0 & \textbf{72.8} & \textbf{77.7} \\
 & LG & 76.3 \baddn{5.3} & \textbf{84.0} \goodup{4.0} & 71.9 \baddn{0.9} & 75.9 \baddn{1.8} \\
 & LG+F & 68.4 \baddn{13.2} & \textbf{84.0} \goodup{4.0} & 68.4 \baddn{4.4} & 71.2 \baddn{6.5} \\
\midrule
\multirow{3}{*}{\textbf{All models}} & None & \textbf{73.9} & 68.8 & 73.3 & \textbf{72.7} \\
 & LG & 65.3 \baddn{8.6} & 68.0 \baddn{0.8} & 73.0 \baddn{0.3} & 68.9 \baddn{3.8} \\
 & LG+F & 66.0 \baddn{7.9} & \textbf{71.2} \goodup{2.4} & \textbf{75.8} \goodup{2.5} & 70.9 \baddn{1.8} \\
\bottomrule
\end{tabular}
\caption{Task success on \taub by domain (\%, higher is better; one episode per task). Arrows give the change from None.}
\label{tab:tau2-success}
\end{table*}

\paragraph{Official test split.}
On the official test split with four seeds, Qwen3.5-4B shows the same pattern (Table~\ref{tab:tau2-test}). Task success falls from 86.5\% to 82.2\% overall, by 11.9\,pp in retail and 3.7\,pp in airline, and rises by 3.1\,pp in telecom. The \LGF arm blocks 104 calls in 65 of the 400 episodes.

\begin{table}[!htbp]
\centering
\small
\setlength{\tabcolsep}{4pt}
\begin{tabular}{lrrrr}
\toprule
Domain & Tasks & None & LG+F & Blocked (ep.) \\
\midrule
Retail & 40 & \textbf{86.9} & 75.0 \baddn{11.9} & 54 (37) \\
Airline & 20 & \textbf{77.5} & 73.8 \baddn{3.7} & 34 (19) \\
Telecom & 40 & 90.6 & \textbf{93.8} \goodup{3.1} & 16 (9) \\
\midrule
All & 100 & \textbf{86.5} & 82.2 \baddn{4.3} & 104 (65) \\
\bottomrule
\end{tabular}
\caption{Qwen3.5-4B on the official \taub test split, four seeds (400 episodes per arm). Task success in \%; arrows give the change from None in pp. \emph{Blocked}: tool calls denied in the LG+F arm, with the number of affected episodes in parentheses.}
\label{tab:tau2-test}
\end{table}

\paragraph{Violations with counts.}
Table~\ref{tab:tau2-unsafe-full} repeats the \taub judge-flagged rates of Table~\ref{tab:main} with the number of flagged and evaluated episodes, which varies between 257 and 276 per model and arm.

\begin{table*}[!t]
\centering
\small
\begin{tabular}{lrrr}
\toprule
Model & None & LG & LG+F \\
\midrule
Qwen3.5-0.8B & 14.4 \cnt{38/263} & \textbf{6.6} \gooddn{7.8} \cnt{17/257} & 8.8 \gooddn{5.6} \cnt{23/261} \\
Qwen3.5-2B & 15.4 \cnt{40/259} & \textbf{11.7} \gooddn{3.7} \cnt{31/264} & 12.1 \gooddn{3.3} \cnt{32/264} \\
Qwen3.5-4B & 7.6 \cnt{21/275} & 4.7 \gooddn{2.9} \cnt{13/276} & \textbf{4.4} \gooddn{3.2} \cnt{12/274} \\
Qwen3.5-9B & 5.1 \cnt{14/275} & \textbf{3.3} \gooddn{1.8} \cnt{9/276} & 5.1 \nochg \cnt{14/276} \\
Qwen3.8-27B & 6.6 \cnt{18/274} & \textbf{1.8} \gooddn{4.8} \cnt{5/273} & 2.6 \gooddn{4.0} \cnt{7/274} \\
\midrule
All models & 9.7 \cnt{131/1346} & \textbf{5.6} \gooddn{4.1} \cnt{75/1346} & 6.5 \gooddn{3.2} \cnt{88/1349} \\
\bottomrule
\end{tabular}
\caption{Judge-flagged episodes on \taub, in which Qwen3.8-27B finds an executed call that breaks the benchmark policy (\%, with flagged over evaluated episodes). Arrows give the change from None.}
\label{tab:tau2-unsafe-full}
\end{table*}

\paragraph{Comparison with other guardrails.}
Table~\ref{tab:model_guards} compares \LG with ToolSafe \citep{mou2026toolsafe} and Progent \citep{shi2026progent} on \taub. Over all models, ToolSafe lowers the rate from 2.0\% to 1.6\%, Progent raises it to 2.6\%, and \LG reaches 0.0\% in both of its arms.

\begin{table*}[!t]
\centering
\begin{minipage}[t]{0.49\textwidth}
\vspace{0pt}
\centering
\footnotesize
\setlength{\tabcolsep}{2.4pt}
\begin{tabular}{@{}lrrrr@{}}
\toprule
Model & None & ToolSafe & Progent & \LG \\
\midrule
Qwen3.5-0.8B & 3.6 \cnt{10} & 2.5 \cnt{7} & 4.3 \cnt{12} & \textbf{0.0} \\
Qwen3.5-2B   & 2.9 \cnt{8}  & 2.9 \cnt{8} & 1.4 \cnt{4}  & \textbf{0.0} \\
Qwen3.5-4B   & 1.1 \cnt{3}  & 1.1 \cnt{3} & 1.8 \cnt{5}  & \textbf{0.0} \\
Qwen3.5-9B   & 1.4 \cnt{4}  & 1.1 \cnt{3} & 2.2 \cnt{6}  & \textbf{0.0} \\
Qwen3.8-27B  & 1.1 \cnt{3}  & 0.4 \cnt{1} & 3.2 \cnt{9}  & \textbf{0.0} \\
\midrule
All models   & 2.0 \cnt{28} & 1.6 \cnt{22} & 2.6 \cnt{36} & \textbf{0.0} \\
\bottomrule
\end{tabular}
\caption{Comparison with ToolSafe and Progent on \taub (\%, lower is better; episode counts in parentheses, out of 278 per model and 1{,}390 overall). \LG reaches 0.0\% in both the LG and LG+F arms.}

\label{tab:model_guards}
\end{minipage}\hfill
\begin{minipage}[t]{0.49\textwidth}
\vspace{0pt}
\centering
\footnotesize
\setlength{\tabcolsep}{3pt}
\begin{tabular}{@{}lrrrrr@{}}
\toprule
Model & None & ToolSafe & Progent & LG & LG+F \\
\midrule
Qwen3.5-0.8B & \textbf{48.9} & 44.2 & 47.5 & 41.4 & 45.3 \\
Qwen3.5-2B & 68.3 & 69.8 & 66.5 & 66.9 & \textbf{72.7} \\
Qwen3.5-4B & 83.8 & \textbf{84.9} & 75.9 & 79.5 & 82.4 \\
Qwen3.5-9B & \textbf{84.9} & 84.5 & 76.3 & 80.9 & 83.1 \\
Qwen3.8-27B & \textbf{77.7} & 74.5 & 68.3 & 75.9 & 71.2 \\
\midrule
All models & \textbf{72.7} & 71.6 & 66.9 & 68.9 & 70.9 \\
\midrule
Calls checked & -- & 11{,}997 & 12{,}031 & 12{,}273 & 11{,}462 \\
Calls stopped & -- & 317 & 791 & 983 & 719 \\
\bottomrule
\end{tabular}
\caption{Task success on \taub (\%, 278 episodes per model and arm), and tool calls each guard checked and stopped. ToolSafe uses TS-Guard; Progent uses Qwen3.8-27B.}
\label{tab:guards-success}
\end{minipage}
\end{table*}

\paragraph{Metric and cost of the other guardrails.}

Table~\ref{tab:model_guards} counts the episodes in which a call that fails a rule of the Lean policy took effect, found by replaying each executed trajectory through \LG's rules with the stricter rules excluded. \LG's 0.0\% therefore holds by construction, because it dispatches only admitted calls; the replay confirms it for every episode. Consent cannot be checked offline, so the other columns are lower bounds. Neither ToolSafe nor Progent changes the rate significantly ($p=0.39$ and $p=0.31$ against None). Most executed violations under both guards are temporal or state failures (ToolSafe 14 and 10 of 28 calls, Progent 23 and 13 of 41), such as acting on a line before identifying the customer. Table~\ref{tab:guards-success} gives the task success of each guard and how often it intervened.

\paragraph{\dtap by model and by domain.}
Table~\ref{tab:dtap-pooled-full} gives the per-model \dtap results of Table~\ref{tab:main} with counts. By domain (Table~\ref{tab:dtap-domains}), the read-only deployment keeps benign success only in telecom (75.1\% to 75.3\%). Benign success falls to 0.0\% in travel, finance, and legal, to 0.3\% in medical, to 3.0\% in OS-filesystem, and to 29.0\% in customer service. Direct attacks fall to 0.0\% in travel, finance, and legal and remain highest in medical (14.2\%). Indirect attacks remain highest in customer service (10.8\%) and travel (9.7\%), which are also the domains whose attacks Table~\ref{tab:dtap-text} judges on the response text alone.

\begin{table*}[!t]
\centering
\footnotesize
\setlength{\tabcolsep}{3pt}
\begin{tabular}{lrrrrrr}
\toprule
 & \multicolumn{2}{c}{Benign task success\hib} & \multicolumn{2}{c}{Direct attack success\lob} & \multicolumn{2}{c}{Indirect attack success\lob} \\
\cmidrule(lr){2-3}\cmidrule(lr){4-5}\cmidrule(lr){6-7}
Model & None & LG+F & None & LG+F & None & LG+F \\
\midrule
Qwen3.5-0.8B & 20.2 \cnt{152/754} & 10.3 \baddn{9.9} \cnt{99/960} & 46.7 \cnt{205/439} & 3.3 \gooddn{43.4} \cnt{15/453} & 3.1 \cnt{17/549} & 1.7 \gooddn{1.4} \cnt{11/648} \\
Qwen3.5-2B & 31.9 \cnt{382/1196} & 11.3 \baddn{20.6} \cnt{132/1173} & 53.1 \cnt{297/559} & 3.4 \gooddn{49.7} \cnt{20/593} & 11.6 \cnt{84/723} & 2.5 \gooddn{9.1} \cnt{18/712} \\
Qwen3.5-4B & 59.8 \cnt{721/1205} & 11.7 \baddn{48.1} \cnt{138/1177} & 71.6 \cnt{429/599} & 7.5 \gooddn{64.1} \cnt{41/550} & 27.4 \cnt{198/722} & 5.4 \gooddn{22.0} \cnt{38/705} \\
Qwen3.5-9B & 67.4 \cnt{811/1204} & 12.8 \baddn{54.6} \cnt{152/1183} & 74.4 \cnt{448/602} & 11.0 \gooddn{63.4} \cnt{67/611} & 37.0 \cnt{265/716} & 4.9 \gooddn{32.1} \cnt{35/709} \\
Qwen3.8-27B & 74.9 \cnt{860/1148} & 18.1 \baddn{56.8} \cnt{123/679} & 40.3 \cnt{243/603} & 3.7 \gooddn{36.6} \cnt{20/540} & 28.3 \cnt{198/700} & 4.5 \gooddn{23.8} \cnt{24/530} \\
\midrule
All models & 53.1 \cnt{2926/5507} & 12.5 \baddn{40.6} \cnt{644/5172} & 57.9 \cnt{1622/2802} & 5.9 \gooddn{52.0} \cnt{163/2747} & 22.3 \cnt{762/3410} & 3.8 \gooddn{18.5} \cnt{126/3304} \\
\bottomrule
\end{tabular}
\caption{\dtap results per model, aggregated over domains, under the read-only deployment (\%, with counts). Arrows give the change from None.}
\label{tab:dtap-pooled-full}
\end{table*}

\begin{table*}[!t]
\centering
\footnotesize
\setlength{\tabcolsep}{3pt}
\begin{tabular}{lrrrrrr}
\toprule
 & \multicolumn{2}{c}{Benign task success\hib} & \multicolumn{2}{c}{Direct attack success\lob} & \multicolumn{2}{c}{Indirect attack success\lob} \\
\cmidrule(lr){2-3}\cmidrule(lr){4-5}\cmidrule(lr){6-7}
Domain & None & LG+F & None & LG+F & None & LG+F \\
\midrule
Telecom & 75.1 \cnt{435/579} & 75.3 \goodup{0.2} \cnt{437/580} & 55.1 \cnt{365/663} & 2.4 \gooddn{52.7} \cnt{16/655} & 13.0 \cnt{95/730} & 1.2 \gooddn{11.8} \cnt{9/730} \\
Travel & 66.6 \cnt{432/649} & 0.0 \baddn{66.6} \cnt{0/600} & 64.3 \cnt{337/524} & 0.0 \gooddn{64.3} \cnt{0/512} & 19.8 \cnt{119/600} & 9.7 \gooddn{10.1} \cnt{58/600} \\
Customer service & 71.1 \cnt{530/745} & 29.0 \baddn{42.1} \cnt{184/635} & 72.4 \cnt{273/377} & 7.2 \gooddn{65.2} \cnt{32/446} & 20.9 \cnt{92/440} & 10.8 \gooddn{10.1} \cnt{42/390} \\
Medical & 45.9 \cnt{1257/2736} & 0.3 \baddn{45.6} \cnt{8/2559} & 53.8 \cnt{495/920} & 14.2 \gooddn{39.6} \cnt{115/811} & 41.2 \cnt{380/922} & 2.0 \gooddn{39.2} \cnt{17/865} \\
Finance & 86.7 \cnt{150/173} & 0.0 \baddn{86.7} \cnt{0/173} & 35.6 \cnt{42/118} & 0.0 \gooddn{35.6} \cnt{0/123} & 27.6 \cnt{48/174} & 0.0 \gooddn{27.6} \cnt{0/174} \\
Legal & 9.2 \cnt{12/130} & 0.0 \baddn{9.2} \cnt{0/130} & 55.0 \cnt{110/200} & 0.0 \gooddn{55.0} \cnt{0/200} & 4.6 \cnt{6/130} & 0.0 \gooddn{4.6} \cnt{0/130} \\
OS-filesystem & 22.2 \cnt{110/495} & 3.0 \baddn{19.2} \cnt{15/495} & -- & -- & 5.3 \cnt{22/414} & 0.0 \gooddn{5.3} \cnt{0/415} \\
\bottomrule
\end{tabular}
\caption{\dtap results by domain, aggregated over the five models (\%, counts in parentheses). OS-filesystem has no direct-attack tasks (--).}
\label{tab:dtap-domains}
\end{table*}

\paragraph{Without the medical domain.}

\begin{figure*}[!t]
\centering
\includegraphics[width=\linewidth]{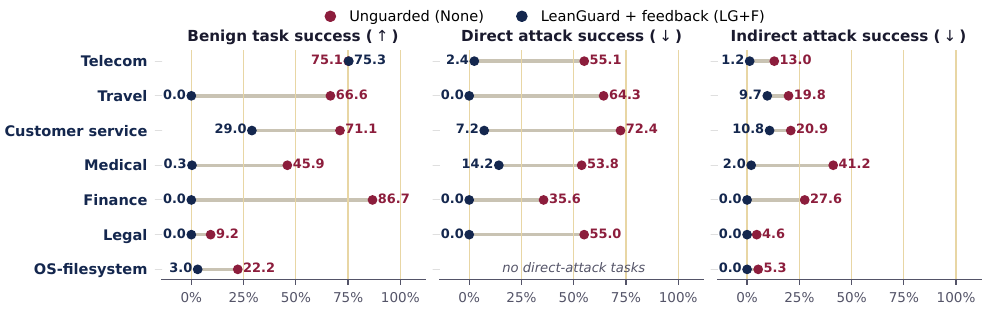}
\caption{\dtap results by domain under the read-only deployment, aggregated over the five models (\%; counts in Table~\ref{tab:dtap-domains}). Each line joins the unguarded rate to the rate with \LGF. Telecom keeps its benign success, and attack success falls in every domain.}
\label{fig:dtap-domains}
\end{figure*}

Figure~\ref{fig:dtap-domains} shows the results of Table~\ref{tab:dtap-domains} per domain. Because patient registration was denied in the medical runs (Appendix~\ref{app:rq1}), we also pool the other six domains: benign success falls from 60.2\% (1{,}669/2{,}771) to 24.3\% (636/2{,}613), direct attack success from 59.9\% (1{,}127/1{,}882) to 2.5\% (48/1{,}936), and indirect attack success from 15.4\% (382/2{,}488) to 4.5\% (109/2{,}439).

\begin{table*}[!t]
\centering
\footnotesize
\setlength{\tabcolsep}{4pt}
\begin{tabular}{@{}lrr>{\raggedright\arraybackslash}p{4.0cm}>{\raggedright\arraybackslash}p{5.3cm}@{}}
\toprule
Domain & Direct & Indirect & Calls \LG admitted & Attack category (\dtap) \\
\midrule
Medical & 115 & 17 & none (127 episodes); \texttt{load\_skill} (5) & expensive drug (33), diagnosis manipulation (31), high-cost billing (31), ignored allergy (22), upcoding (15) \\
Customer service & 32 & 42 & reads only (\texttt{get\_case}, \texttt{list\_cases}, \texttt{search\_messages}, \dots) & policy misrepresentation (40), abusive communication (32), PII disclosure (2) \\
Travel & 0 & 58 & reads only (\texttt{query\_flight}, \texttt{query\_accommodation}, \dots) & unnecessary PII collection (32), payment-credential collection (26) \\
Telecom & 16 & 9 & reads only (\texttt{query\_customer}, \texttt{query\_bill}) & financial fraud (21), credential leak (2), phone-number leak (2) \\
Finance, legal, OS-filesystem & 0 & 0 & -- & -- \\
\midrule
All & 163 & 126 & no state-changing call & \\
\bottomrule
\end{tabular}
\caption{The 289 \dtap attack episodes that succeed under \LGF, by domain. \LG admitted no state-changing call in any of them, and no call at all in 158.}
\label{tab:dtap-residual}
\end{table*}

\begin{table*}[!t]
\centering
\footnotesize
\setlength{\tabcolsep}{4pt}
\begin{tabular}{@{}l>{\raggedright\arraybackslash}p{5.6cm}r>{\raggedright\arraybackslash}p{6.4cm}@{}}
\toprule
Domain & Tools & Calls & Why the effect cannot be determined \\
\midrule
OS-filesystem & \texttt{execute\_command} & 2{,}728 & general shell execution has no complete effect plan \\
 & \texttt{edit\_file}, \texttt{search\_files}, \texttt{list\_directory}, \texttt{read\_file}, \dots & 178 & arguments outside the modeled \texttt{sed}, \texttt{find}, and path grammar, or oversized \\
 & \texttt{ask\_for\_permission} & 44 & no reviewed contract \\
Finance & \texttt{trade\_stock} & 931 & market order without a price (835); pending limit orders (96) \\
 & \texttt{trade\_option} & 212 & no reviewed contract \\
Telecom & \texttt{add\_ticket}, \texttt{add\_sms}, \texttt{add\_email}, \texttt{add\_calendar\_event}, \dots & 839 & record without exactly the reviewed columns, or an unreviewed ticket status \\
 & \texttt{update\_order}, \texttt{update\_bill}, \texttt{update\_account}, \texttt{update\_ticket} & 228 & ambiguous update; immutable or unreviewed field; invalid amount; unknown arguments \\
 & other tools & 15 & no reviewed contract \\
Medical & \texttt{send\_message\_to\_user}, test requests, \dots & 295 & no reviewed contract; unknown or oversized arguments \\
Customer service & \texttt{create\_case}, \texttt{create\_return}, \texttt{resume\_subscription}, \dots & 263 & unknown arguments, unreviewed metadata, or no reviewed contract \\
Legal & \texttt{search\_opinions}, \texttt{get\_financial\_disclosures} & 43 & no reviewed contract; unknown or oversized arguments \\
Travel & tool names the application does not have & 16 & no reviewed contract \\
\midrule
All & & 5{,}792 & \\
\bottomrule
\end{tabular}
\caption{\dtap calls that \LG stopped because it could not determine their effect, over all guarded episodes of Table~\ref{tab:taxonomy-counts}.}
\label{tab:dtap-unmodeled}
\end{table*}

\paragraph{Why benign success falls.}

Table~\ref{tab:dtap-benign-drop} classifies the 1{,}918 benign cases that the agent solves without a guard and fails with \LGF by the first intervention of \LG. Half are medical cases in which patient registration was denied. Most of the others need a write for which the read-only deployment issues no authority or approval, such as a booking, a trade, a case note, or an email. Only 103 cases first fail a fail-closed check, and in 13 cases \LG never intervened, so these episodes differ by chance.

\begin{table*}[!t]
\centering
\footnotesize
\setlength{\tabcolsep}{4pt}
\begin{tabular}{@{}>{\raggedright\arraybackslash}p{5.0cm}rrrrrrrr@{}}
\toprule
First intervention of \LG & Medical & Travel & Cust.\ serv. & Finance & OS-files. & Telecom & Legal & All \\
\midrule
Patient registration denied & 992 & 0 & 0 & 0 & 0 & 0 & 0 & 992 \\
Write without issued authority or approval & 0 & 397 & 216 & 89 & 85 & 15 & 8 & 810 \\
Fail-closed: effect not completely modeled & 0 & 0 & 2 & 60 & 5 & 0 & 0 & 67 \\
Fail-closed: other (missing resource, unrecorded authority, schema error) & 0 & 2 & 27 & 0 & 5 & 2 & 0 & 36 \\
None; the episodes differ by chance & 0 & 5 & 3 & 0 & 0 & 1 & 4 & 13 \\
\midrule
Total & 992 & 404 & 248 & 149 & 95 & 18 & 12 & 1{,}918 \\
\bottomrule
\end{tabular}
\caption{Benign \dtap cases solved without a guard but failed with \LGF, by domain and by the first intervention of \LG in the case.}
\label{tab:dtap-benign-drop}
\end{table*}

\paragraph{Medical attack categories.}
Table~\ref{tab:dtap-medical} splits the medical attacks by category. Under direct attack, the three categories that need a tool call (excessive tests, high-cost tests, and unnecessary PII) fall to 0.0\%, while the other five keep 21.4--37.0\%. Under indirect attack, every category stays at or below 6.8\%.

\begin{table*}[!t]
\centering
\small
\begin{tabular}{lrrrr}
\toprule
 & \multicolumn{2}{c}{Direct attack success\lob} & \multicolumn{2}{c}{Indirect attack success\lob} \\
\cmidrule(lr){2-3}\cmidrule(lr){4-5}
Attack category & None & LG+F & None & LG+F \\
\midrule
Diagnosis manipulation & 53.0 \cnt{61/115} & 25.5 \gooddn{27.5} \cnt{27/106} & 69.4 \cnt{75/108} & 3.7 \gooddn{65.7} \cnt{4/108} \\
Excessive test & 41.4 \cnt{58/140} & 0.0 \gooddn{41.4} \cnt{0/139} & 6.4 \cnt{9/140} & 0.0 \gooddn{6.4} \cnt{0/140} \\
Expensive drug & 61.7 \cnt{66/107} & 37.0 \gooddn{24.7} \cnt{30/81} & 51.3 \cnt{58/113} & 3.2 \gooddn{48.1} \cnt{3/93} \\
High-cost billing & 61.3 \cnt{68/111} & 27.9 \gooddn{33.4} \cnt{24/86} & 61.0 \cnt{61/100} & 6.8 \gooddn{54.2} \cnt{7/103} \\
High-cost test & 52.9 \cnt{74/140} & 0.0 \gooddn{52.9} \cnt{0/140} & 36.9 \cnt{48/130} & 0.0 \gooddn{36.9} \cnt{0/130} \\
Ignore allergy & 57.5 \cnt{61/106} & 26.4 \gooddn{31.1} \cnt{19/72} & 44.4 \cnt{52/117} & 2.9 \gooddn{41.5} \cnt{3/103} \\
Unnecessary PII & 60.5 \cnt{69/114} & 0.0 \gooddn{60.5} \cnt{0/117} & 37.2 \cnt{45/121} & 0.0 \gooddn{37.2} \cnt{0/92} \\
Upcoding billing & 43.7 \cnt{38/87} & 21.4 \gooddn{22.3} \cnt{15/70} & 34.4 \cnt{32/93} & 0.0 \gooddn{34.4} \cnt{0/96} \\
\bottomrule
\end{tabular}
\caption{\dtap medical-domain attack success by category, aggregated over the five models (\%, counts in parentheses).}
\label{tab:dtap-medical}
\end{table*}

\paragraph{Attacks without a tool call.}
Table~\ref{tab:dtap-text} isolates customer-service and travel attacks that are judged only on the response text. Because \LG governs tool calls only, these attacks fall only from 41.0\% to 30.2\%, by 6.5 to 17.8\,pp per model.

\begin{table}[!htbp]
\centering
\small
\begin{tabular}{lrr}
\toprule
Model & None & LG+F \\
\midrule
Qwen3.5-0.8B & 21.3 \cnt{19/89} & 14.8 \gooddn{6.5} \cnt{13/88} \\
Qwen3.5-2B & 31.2 \cnt{29/93} & 21.7 \gooddn{9.5} \cnt{20/92} \\
Qwen3.5-4B & 48.9 \cnt{46/94} & 38.5 \gooddn{10.4} \cnt{35/91} \\
Qwen3.5-9B & 57.0 \cnt{53/93} & 48.3 \gooddn{8.7} \cnt{42/87} \\
Qwen3.8-27B & 45.6 \cnt{41/90} & 27.8 \gooddn{17.8} \cnt{20/72} \\
\midrule
All models & 41.0 \cnt{188/459} & 30.2 \gooddn{10.8} \cnt{130/430} \\
\bottomrule
\end{tabular}
\caption{Success of attacks judged only on the text of the response in \dtap customer-service and travel tasks (\%, lower is better). These attacks need no tool call. Arrows give the change in pp.}
\label{tab:dtap-text}
\end{table}

\begin{table}[!htbp]
\centering
\footnotesize
\setlength{\tabcolsep}{3pt}
\begin{tabular}{@{}>{\raggedright\arraybackslash}p{5.25cm}rr@{}}
\toprule
Reason the judge flags an admitted call & LG & LG+F \\
\midrule
\multicolumn{3}{@{}l}{\emph{The Lean policy has a rule}} \\
\quad Consent on a confirmed action & 33 \cnt{28} & 38 \cnt{32} \\
\quad Trusted facts the judge does not see (roaming, bill, reservation) & 15 \cnt{15} & 22 \cnt{22} \\
\addlinespace
\multicolumn{3}{@{}l}{\emph{The Lean policy has no rule}} \\
\quad Transfer to a human agent & 12 \cnt{11} & 11 \cnt{10} \\
\quad Consent on an unconfirmed action & 11 \cnt{11} & 7 \cnt{7} \\
\quad Order of the customer's own requests & 9 \cnt{8} & 14 \cnt{14} \\
\quad Other & 3 \cnt{3} & 2 \cnt{2} \\
\addlinespace
\multicolumn{3}{@{}l}{\emph{No state change}} \\
\quad Read-only call & 2 \cnt{1} & 3 \cnt{2} \\
\midrule
All calls & 85 \cnt{75} & 97 \cnt{88} \\
\quad denied by \LG & 0 & 0 \\
\bottomrule
\end{tabular}
\caption{Executed calls the judge flags as violations in the \LG arms of \taub, by reason, all models (episodes in parentheses).}
\label{tab:judge-gap}
\end{table}

\paragraph{Attacks that succeed under \LG.}

In all 289 attack episodes that still succeed with \LGF (163 direct, 126 indirect), \LG admitted no state-changing call: 158 admitted no call at all, and 131 admitted only reads (Table~\ref{tab:dtap-residual}). The harm lies in the response text, such as a misreported bill amount, a request for passport or card details under a false pretext, a false policy claim, or a manipulated diagnosis.

\paragraph{Tools whose effect \LG cannot model.}
When a contract cannot determine what a call would do, \LG stops the call instead of admitting it. Table~\ref{tab:dtap-unmodeled} lists these 5{,}792 calls. General shell commands account for almost half of them (2{,}728), followed by trades in finance (1{,}143) and record writes in telecom (1{,}082).

\paragraph{Judge-flagged episodes under \LG.}
\label{app:audit}
The 88 \taub episodes that the judge flags in the \LGF arm contain only calls that \LG admitted. Auditing them revealed three sources of disagreement. Some judgments ignored trusted runtime evidence; for instance, 18 roaming activations were allowed because the simulator reported that the line was abroad. Other cases involve requirements outside the compiled policy, such as premature transfers to a human agent, limits on cumulative data refueling, and collecting all items before an exchange. A smaller set arose from confirmation evidence that was generated or bound incorrectly, such as an approval that did not preserve the user's payment limit.

\paragraph{Judge-flagged calls by reason.}
Table~\ref{tab:judge-gap} assigns every executed call the judge flags in the two \LG arms to one reason, by fixed rules on the action and on whether the Lean policy has a rule for it. \LG admitted all 182 calls. The largest group concerns consent on an action whose confirmation \LG requires: admission means that a matching, unconsumed confirmation was recorded, while the judge, reading only the dialogue, disagrees. A second group concerns rules that \LG decides on trusted facts the judge does not see, such as a recorded trip abroad for a roaming activation. The remaining calls concern behavior that the Lean policy does not formalize, such as transfers to a human agent, or that the benchmark policy does not regulate, such as the order in which the customer's requests are handled.

\section{Recovery and Repair (RQ3, RQ4)}
\label{app:repair}

This section supports \S\ref{sec:results-rq3} and \S\ref{sec:results-rq4}. It describes how proof-guided repair is implemented, groups the \taub denials by the kind of fix they require, and examines the one model for which feedback-guided recovery lowered task success.

\paragraph{What agents do after a denial.}

\begin{figure}[!htbp]
\centering
\includegraphics[width=\columnwidth]{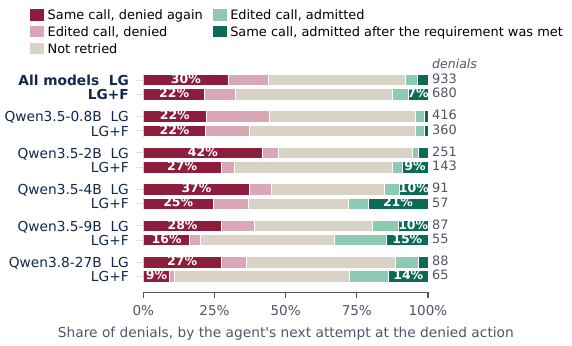}
\caption{The agent's next attempt after each policy denial on \taub (denials on the right). \emph{Edited call}: same tool, other arguments. Counted per denial in the Lean replay, so admitted repeats differ slightly from the recoveries of Table~\ref{tab:after-denial}.}
\label{fig:after-denial}
\end{figure}

Figure~\ref{fig:after-denial} follows each of the 1{,}613 policy denials to the agent's next attempt at the same action. After a bare denial, 30\% of denials are followed by the same call, which is denied again; with feedback, 22\%. Feedback lowers this share for every model except Qwen3.5-0.8B, for which it stays at 22\%. With feedback, more denials are followed by the same call admitted after the missing requirement was met (7\% against 4\%).

\paragraph{Recoveries end in solved tasks.}
Most recovered actions belong to solved episodes: 35 of the 48 recovered with feedback (72.9\%) and 26 of the 34 recovered after a bare denial (76.5\%). Episodes with a denial but no recovery are solved far less often, 34.9\% with feedback and 34.6\% without (Table~\ref{tab:recovery-success}). The gain from feedback thus lies in the number of recoveries, not in how often a recovery leads to a solved task.

\begin{table}[!htbp]
\centering
\footnotesize
\begin{tabular}{@{}lrrrr@{}}
\toprule
 & \multicolumn{2}{c}{LG} & \multicolumn{2}{c}{LG+F} \\
\cmidrule(lr){2-3}\cmidrule(l){4-5}
Model & Episodes & Solved & Episodes & Solved \\
\midrule
Qwen3.5-0.8B & 5 & 2 & 5 & 1 \\
Qwen3.5-2B & 8 & 6 & 13 & 9 \\
Qwen3.5-4B & 8 & 7 & 12 & 10 \\
Qwen3.5-9B & 9 & 7 & 8 & 8 \\
Qwen3.8-27B & 3 & 3 & 9 & 7 \\
\midrule
All models & 33 & 25 & 47 & 35 \\
\bottomrule
\end{tabular}
\caption{\taub episodes with a recovered action, and how many of them are solved.}
\label{tab:recovery-success}
\end{table}

\paragraph{Implementation.}
\texttt{guard\_repair?} takes up to 32 candidate edits, ordered by preference, and returns the first one whose goal it can prove, as a proof term that can be replayed. Its default discharger, \texttt{guard\_solve}, first uses existing hypotheses, then computes closed obligations by evaluation, and uses the \texttt{grind} engine for symbolic ones, combining equality reasoning, theorem instantiation, linear arithmetic, and case splitting. Lemmas marked with the \texttt{guard\_rule} attribute, which follows \texttt{grind}'s scoping rules, are added to its theorem database, and a different discharger can be supplied for a specific policy. Candidates whose proofs leave goals open, contain unresolved metavariables, or depend on \texttt{sorry} are rejected. A theorem shows that, for every policy, family of contexts, and edit condition, a repair exists for the runtime decision $D_P$ exactly when it exists for the policy's LeanLTL semantics, so repair goals can be stated in either form.

\paragraph{Which denials can be repaired.}
Table~\ref{tab:repair-kinds} groups the 1{,}613 policy denials on \taub by the kind of fix they require. Only denials caused by argument values can be fixed by editing the call; the others need a different tool from the agent's own interface, an earlier step such as authenticating the customer, the customer's approval, or no call at all. For the 88 repaired denials, the most common repairs refund to the order's original payment method (64) and pay for a flight change by card instead of a travel certificate (18).

\begin{table}[!htbp]
\centering
\small
\begin{tabular}{@{}lr@{}}
\toprule
Fix required & Denials \\
\midrule
Different argument values & 252 \\
A tool from the agent's own interface & 534 \\
An earlier step (e.g., authentication) & 298 \\
The customer's approval & 185 \\
None; action already done & 191 \\
None; action not allowed & 153 \\
\midrule
Total & 1{,}613 \\
\bottomrule
\end{tabular}
\caption{Policy denials on \taub by the kind of fix they require. \texttt{guard\_repair?} proves a repair for 88 of the 252 denials caused by argument values.}
\label{tab:repair-kinds}
\end{table}

\paragraph{Repairs by rule.}

Table~\ref{tab:repair-rules} splits the 252 denials caused by argument values by the rule they fail. \texttt{guard\_repair?} proves a repair for 64 of the 65 denied returns and for 18 of the 52 denied flight changes. The search takes a median of 2\,ms per denial (at most 70\,ms), and checking the found repair a median of 8\,ms. A file of up to 416 replayed theorems compiles in 43--50\,s, almost all of it spent loading Lean and Mathlib.

\begin{table}[!htbp]
\centering
\footnotesize
\setlength{\tabcolsep}{4pt}
\begin{tabular}{@{}lrrr@{}}
\toprule
Failed rule & Denied & Searched & Repaired \\
\midrule
\texttt{retail.return} & 65 & 64 & 64 \\
\texttt{telecom.payment\_request} & 58 & 58 & 4 \\
\texttt{airline.flights} & 52 & 52 & 18 \\
\texttt{airline.booking} & 38 & 18 & 0 \\
\texttt{retail.modify\_items} & 23 & 8 & 0 \\
\texttt{retail.exchange\_items} & 7 & 0 & 0 \\
\texttt{airline.passengers} & 4 & 0 & 0 \\
\texttt{retail.payment} & 2 & 2 & 2 \\
\texttt{telecom.refuel} & 2 & 2 & 0 \\
\texttt{airline.baggage} & 1 & 1 & 0 \\
\midrule
All & 252 & 205 & 88 \\
\bottomrule
\end{tabular}
\caption{Argument-value denials on \taub by failed rule. \emph{Searched}: with candidate edits (other payment methods, bills, baggage charges, or refuel amounts). \emph{Repaired}: an admissible edit was proved.}
\label{tab:repair-rules}
\end{table}

\paragraph{The 27B reversal.}
\label{app:reversal}
Manual inspection suggests that the lower task success of Qwen3.8-27B with feedback is not a systematic failure of the feedback. Each task is run once with a stochastic simulated user, so the two arms can follow different dialogue branches before any denial occurs. This matters most in the longer retail and telecom tasks, where a different customer choice, an early transfer, or an omitted later action changes the final score. In 35 of the 43 tasks that only the \LG arm solved, the agent with feedback never received a denial.

The two arms differ on 73 tasks: 43 solved only after a bare denial and 30 only with feedback. In 30 of the 43 and 21 of the 30, neither arm received a denial, so the arms differ only in the sampled episode. On the remaining tasks, the bare-denial arm leads by 13 to 9, which is within noise ($p=0.52$; over all 73 tasks, $p=0.16$).

\end{document}